%% file: Formatting-Instructions-LaTeX-2024.tex
\documentclass[letterpaper]{article} 
\usepackage{aaai24}  
\usepackage{times}  
\usepackage{helvet}  
\usepackage{courier}  
\usepackage[hyphens]{url}  
\usepackage{graphicx} 
\usepackage{natbib}  
\usepackage{caption} 
\usepackage{algorithm}
\usepackage{algorithmic}

\usepackage{lipsum,booktabs}
\usepackage{amsmath,amsfonts,amssymb}
\usepackage{amsthm}
\usepackage{graphicx}
\usepackage{color}
\usepackage{diagbox}
\usepackage{makecell}
\usepackage{multicol}
\usepackage{multirow}
\usepackage{tabularx}
\usepackage{paralist}
\usepackage{array}
\usepackage{graphicx} 

\usepackage{listings}
\usepackage{color}

\usepackage[capitalize,noabbrev]{cleveref}

\usepackage{listings}
\usepackage{color}

\definecolor{dkgreen}{rgb}{0,0.6,0}
\definecolor{gray}{rgb}{0.5,0.5,0.5}
\definecolor{mauve}{rgb}{0.58,0,0.82}

\usepackage{newfloat}
\usepackage{listings}
\DeclareCaptionStyle{ruled}{labelfont=normalfont,labelsep=colon,strut=off} 
\floatstyle{ruled}
\newfloat{listing}{tb}{lst}{}
\floatname{listing}{Listing}
\def\eg{\emph{e.g.,}}

\def\ie{\emph{i.e., }}

\newcommand*{\tran}{^{\mkern-1.5mu\mathsf{T}}}
\newcommand{\eq}{Eq.}

\title{Exploring Transformer Extrapolation}
\author {
    \textsuperscript{\rm 1}Zhen Qin\equalcontrib,
    \textsuperscript{\rm 1}Yiran Zhong\equalcontrib
    \thanks{Corresponding author. Email: \texttt{zhongyiran@gmail.com}.},
    \textsuperscript{\rm 2}Hui Deng
}
\affiliations {
    \textsuperscript{\rm 1}OpenNLPLab, Shanghai Artificial Intelligence Laboratory,
    \textsuperscript{\rm 2}Northwestern Polytechnical University\\
    https://github.com/OpenNLPLab/Rpe
}

\usepackage{bibentry}

\begin{document}

\maketitle

\begin{abstract}
Length extrapolation has attracted considerable attention recently since it allows transformers to be tested on longer sequences than those used in training. Previous research has shown that this property can be attained by using carefully designed Relative Positional Encodings (RPEs). 
While these methods perform well on a variety of corpora, the conditions for length extrapolation have yet to be investigated. 
This paper attempts to determine what types of RPEs allow for length extrapolation through a thorough mathematical and empirical analysis.
We discover that a transformer is certain to possess this property as long as the series that corresponds to the RPE's exponential converges.
Two practices are derived from the conditions and examined in language modeling tasks on a variety of corpora.
As a bonus from the conditions, we derive a new Theoretical Receptive Field (TRF) to measure the receptive field of RPEs without taking any training steps. 
Extensive experiments are conducted on the Wikitext-103, Books, Github, and WikiBook datasets to demonstrate the viability of our discovered conditions. We also compare TRF to Empirical Receptive Field (ERF) across different models, showing consistently matched trends on these datasets. 
\end{abstract}

\section{Introduction}
Transformer~\cite{vaswani2017attention} is advancing steadily in the areas of natural language processing~\cite{qin2023scaling,devlin-etal-2019-bert,liu2019roberta,zhen2022cosformer,qin-etal-2022-devil,liu2022neural,qin-etal-2023-linear}, computer vision~\cite{vit,2206.10552,lu2022linear,hao2023improving}, and audio processing~\cite{gong21b_interspeech,akbari2021vatt,49414,sun2022locality}. Although it outperforms other architectures such as RNNs~\cite{cho-etal-2014-learning,qin2023hierarchically} and CNNs~\cite{kim2014cnn, Hershey2016CNNAF,gehring2017convolutional} in many sequence modeling tasks, its lack of length extrapolation capability limits its ability to handle a wide range of sequence lengths, \ie inference sequences need to be equal to or shorter than training sequences.
Increasing the training sequence length is only a temporary solution because the space-time complexity grows quadratically with the sequence length. Another option is to extend the inference sequence length by converting the trained full attention blocks to sliding window attention blocks~\cite{2004.05150}, but this will result in significantly worse efficiency than the full attention speed~\cite{alibi}. 
How to permanently resolve this issue without incurring additional costs has emerged as a new topic.

A mainstream solution for length extrapolation is to design a Relative Positional Encoding (RPE)~\cite{qin2023linearized} that concentrates attention on neighboring tokens. For example, ALiBi~\cite{alibi} applies linear decay biases to the attention to reduce the contribution from distant tokens. Kerple~\cite{chi2022kerple} investigates shift-invariant conditionally positive definite kernels in RPEs and proposes a collection of kernels that promote the length extrapolation property. It also shows that ALiBi is one of its instances. Sandwich~\cite{chi2022sandwitch} proposes a hypothesis to explain the secret behind ALiBi and empirically proves it by integrating the hypothesis into sinusoidal positional embeddings.

In order to investigate transformer extrapolation, we first establish a hypothesis regarding why existing RPE-based length extrapolation methods~\cite{qin2023toeplitz} have this capacity to extrapolate sequences in inference based on empirical analysis. Then we identify the conditions of RPEs that satisfy the hypothesis through mathematical analysis. Finally, the discovered conditions are empirically validated on a variety of corpora. Specifically, we assume that due to decay biases, existing RPE-based length extrapolation methods behave similarly to sliding window attention, \ie only tokens within a certain range can influence the attention scores. A transformer can extrapolate for certain in this scenario since the out-of-range tokens have no effect on the attention outcomes. We derive that a transformer is guaranteed to satisfy this hypothesis if the series corresponding to the exponential of its RPE converges. Based on the observation, we show that previous RPE-based methods~\cite{alibi,chi2022kerple} can be seen as particular instances under the conditions. Two new practices from the conditions are derived and evaluated in language modeling.

The observed conditions not only shed light on the secret of length extrapolation but also offer a new perspective on computing the Theoretical Receptive Fields (TRF) of RPEs. 
In contrast to prior approaches that require training gradients to compute TRF, we propose a new way to calculate TRF that is solely based on the formulation of RPEs. Extensive experiments on Wikitext-103~\cite{1609.07843}, Books~\cite{Zhu_2015_ICCV}, Github~\cite{pile}, and WikiBook~\cite{2202.08005} validate the conditions. TRF calculated by our method substantially matches the trend of the Empirical Receptive Field (ERF) in real-world scenarios.

\section{Preliminary}
Before embarking on the journey of exploring, we introduce several preliminary concepts that will be used throughout the paper, such as softmax attention, relative positional encoding, length extrapolation, and sliding window attention. We also provide the necessary notations for the subsequent analysis, \ie we use $\mathbf M$ to denote a matrix and $\mathbf m_i^{\top}$ to represent the $i$th row of $\mathbf M$. The complete math notations can be found in Appendix. Following previous work~\cite{alibi}, we restrict our analysis to causal language models and assume that the max sequence length during training is $m$.

\subsection{Softmax attention}
Softmax attention is a key component of transformers which operates on query $\mathbf{Q}$, key $\mathbf{K}$ and value $\mathbf{V}$ matrices. Each matrix is a linear map that takes $\mathbf X\in \mathbb R^{n\times d}$ as input:
\small 
\begin{equation}
    \mathbf{Q} = \mathbf{X}\mathbf{W}_Q , \
 \mathbf{K} =\mathbf{X}\mathbf{W}_K, \
\mathbf{V} =\mathbf{X}\mathbf{W}_V \in \mathbb{R}^{n\times d},
\end{equation}
\normalsize
where $n$ is the sequence length and $d$ is the dimension of the hidden feature.
The output attention matrix $\mathbf{O}\in \mathbb R^{n\times d}$ can be formulated as:
\small 
\begin{equation}
\label{att}
\mathbf{O}= \mathrm{Softmax}(\mathbf{Q} \mathbf{K}\tran / \sqrt{d}) \mathbf{V}.
\end{equation}
\normalsize
To prevent information leakage in causal language modeling, a mask matrix $\mathbf M\in \mathbb R^{n\times n}$ is used to ensure that current tokens can only see previous tokens and themselves. The lower triangular elements of $\mathbf M$ are $0$, and the upper triangular ones, except for the diagonal, are $-\infty$. Then the output attention matrix $\mathbf{O}$ for causal language models will be:
\small 
\begin{equation}
\label{att_mask}
\mathbf O= \mathrm{Softmax}(\mathbf Q\mathbf K^{\top}/\sqrt{d}+\mathbf M)\mathbf V.
\end{equation}
\normalsize
Note that \eq~\ref{att_mask} can be seen as a general form of attention, \ie when the elements of $\mathbf M$ are all 0, \eq~\ref{att_mask} is degenerated to \eq~\ref{att}. For ease of discussion, we use \eq~\ref{att_mask} to represent attention computation.

\subsection{Relative positional encoding}
Positional encoding is designed to inject positional bias into transformers. Absolute Positional Encoding (APE)~\cite{vaswani2017attention,gehring2017convolutional} and Relative Positional Encoding (RPE)~\cite{su2021roformer,liutkus2021relative,alibi,chi2022kerple} are the two most common types of positional encoding. In this paper, we focus on RPE because it is the key for length extrapolation, as shown in~\cite{alibi}.
An attention with RPE can be written as:
\small 
\begin{equation}
\label{att_rpeo}
\mathbf O= \mathrm{Softmax}(\mathbf Q\mathbf K^{\top}/\sqrt{d} +\mathbf M+\mathbf P)\mathbf V,
\end{equation}
\normalsize
where $\mathbf P\in \mathbb R^{n\times n}$ is a Toeplitz matrix that encodes relative positional information, \ie $ p_{ij}=p_{i-j}$. It is worth noting that $\mathbf M$ and $\mathbf P$ can be merged, and the merged matrix is still a Toeplitz matrix. We use $\mathbf R$ to represent the merged matrix and rewrite \eq~\ref{att_rpeo} as:
\small  
\begin{equation}
\label{att_rpe}
\mathbf O= \mathrm{Softmax}(\mathbf Q\mathbf K^{\top}/\sqrt{d} +\mathbf R)\mathbf V.
\end{equation}
\normalsize

\subsection{Definition of length extrapolation}
The property of length extrapolation allows a model to be tested on longer sequences than those used in training. Previous sequence modeling structures such as RNNs~\cite{hochreiter1997long} and CNNs~\cite{gehring2017convolutional} often naturally possess this property, but it is a difficult task for transformers. This property is only present in sliding window transformers and a few transformer variants with specifically designed RPEs~\cite{chi2022kerple,alibi,chi2022sandwitch}.

In language modeling, one token can only see itself and previous tokens. Therefore, regardless the sequence length, the performance should be stable for the neighboring tokens that are within the training sequence length~\cite{2004.05150}. For the tokens that are out of range, the performance will degrade if the model does not support length extrapolation~\cite{alibi}. Based on the observation above, we give a definition of length extrapolation:

\newtheorem{extrap}{Definition}[section]
\begin{extrap}
\label{extrap}
For a language model $\mathcal F$, given dataset $\mathcal X$, if for any $n$, there is, 
\small
\begin{equation}
|ppl_n(\mathcal X, \mathcal F)- ppl_m(\mathcal X,\mathcal F)|/ppl_m(\mathcal X,\mathcal F) < \delta,
\end{equation}
 \normalsize
then $\mathcal F$ is considered to have the extrapolation property. 
\end{extrap}
Here $\delta>0$ is a small constant, $ppl_n(\mathcal X, \mathcal F)$ means that $\mathcal F$ calculates perplexity with a max sequence length of $n$ on the data set $\mathcal X$.
Empirically, if $|ppl_n(\mathcal X, \mathcal F)- ppl_m(\mathcal X,\mathcal F)|/ppl_m(\mathcal X,\mathcal F)$  becomes very large($\gg 1$) as $n$ increases, we consider that $\mathcal F$ does not have extrapolation property.

\subsection{Sliding window attention}
For the convenience of subsequent discussions, we define a window attention at position $i$ and window size $j$ as follows: 
\small
\begin{equation}
\label{att_trun}
\begin{aligned}
\mathbf o_i^{j}
&= \frac{\sum_{i -j + 1\le s\le i}\exp(\mathbf q_i^{\top}\mathbf k_s/\sqrt d )\exp(r_{is})\mathbf v_s}
{\sum_{i -j + 1\le t\le i} \exp(\mathbf q_i^{\top}\mathbf k_t /\sqrt d)\exp(r_{it})} \\
&\triangleq 
\frac{\sum_{i-j+1\le  s\le i}c_{is}\mathbf  v_s}
{C_{ij} },
\end{aligned}
\end{equation}
 \normalsize
where
$
C_{ij} =\sum_{i -j + 1\le t\le i} c_{it}, c_{ij} = a_{ij}  b_{ij}, 
a_{ij} =\exp( \mathbf q_i^{\top}\mathbf k_j/\sqrt d),
b_{ij} =\exp(r_{ij}), 
j\le i.
$

We further assume $ \|\mathbf x_i\|\le l, \mathbf x \in \{\mathbf q, \mathbf k, \mathbf v\}$, where $l>0$ is a constant.
The $\mathbf o_i^j$ represents the attention output of the $i$-th token, which interacts with the $j$ tokens preceding it. Note that window attention naturally possesses the length extrapolation ability.
\begin{figure*}[ht]
\centering
\vspace{-5mm}
\includegraphics[width=0.48\linewidth]{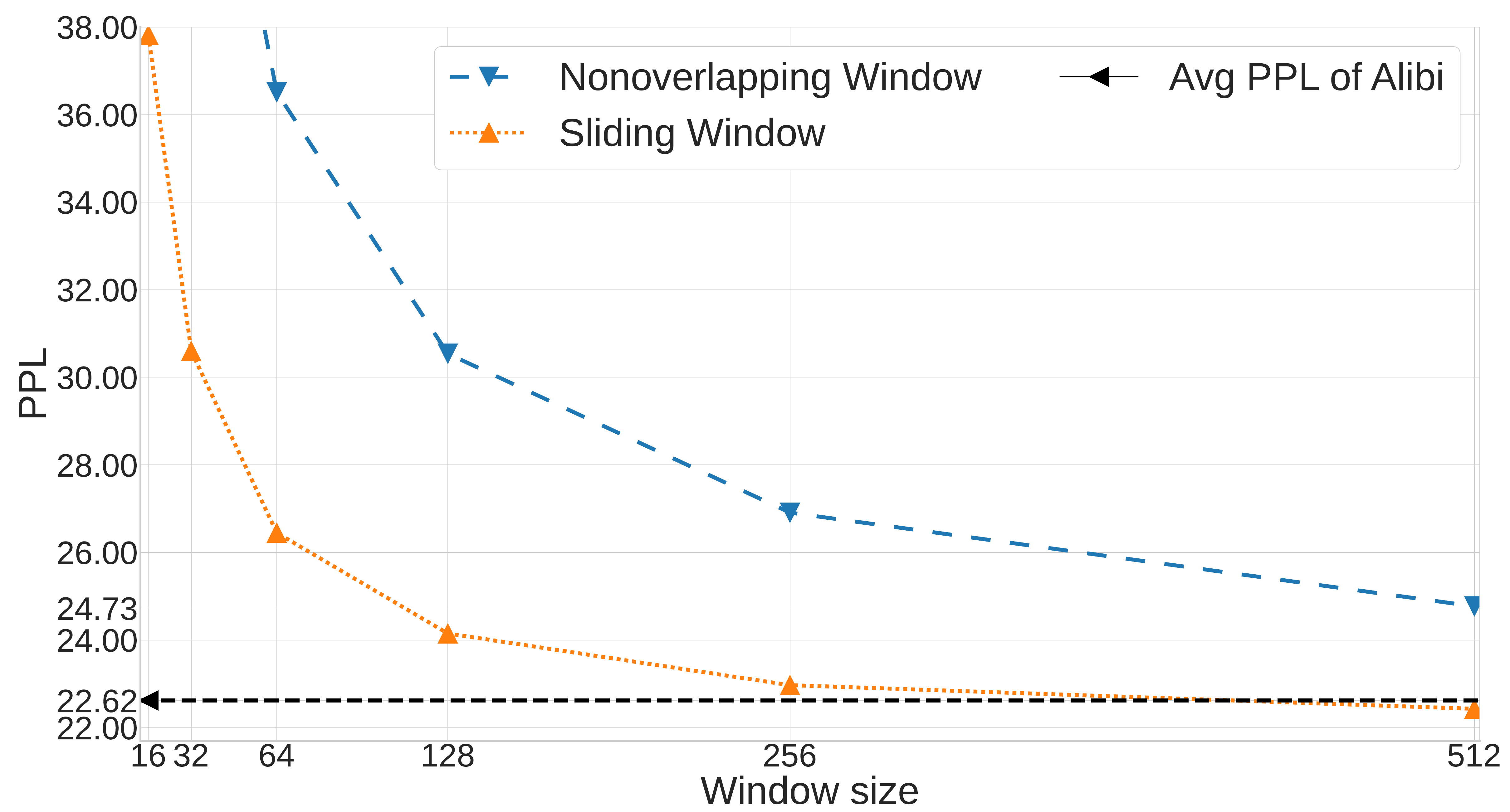}  
\includegraphics[width=0.48\linewidth]
{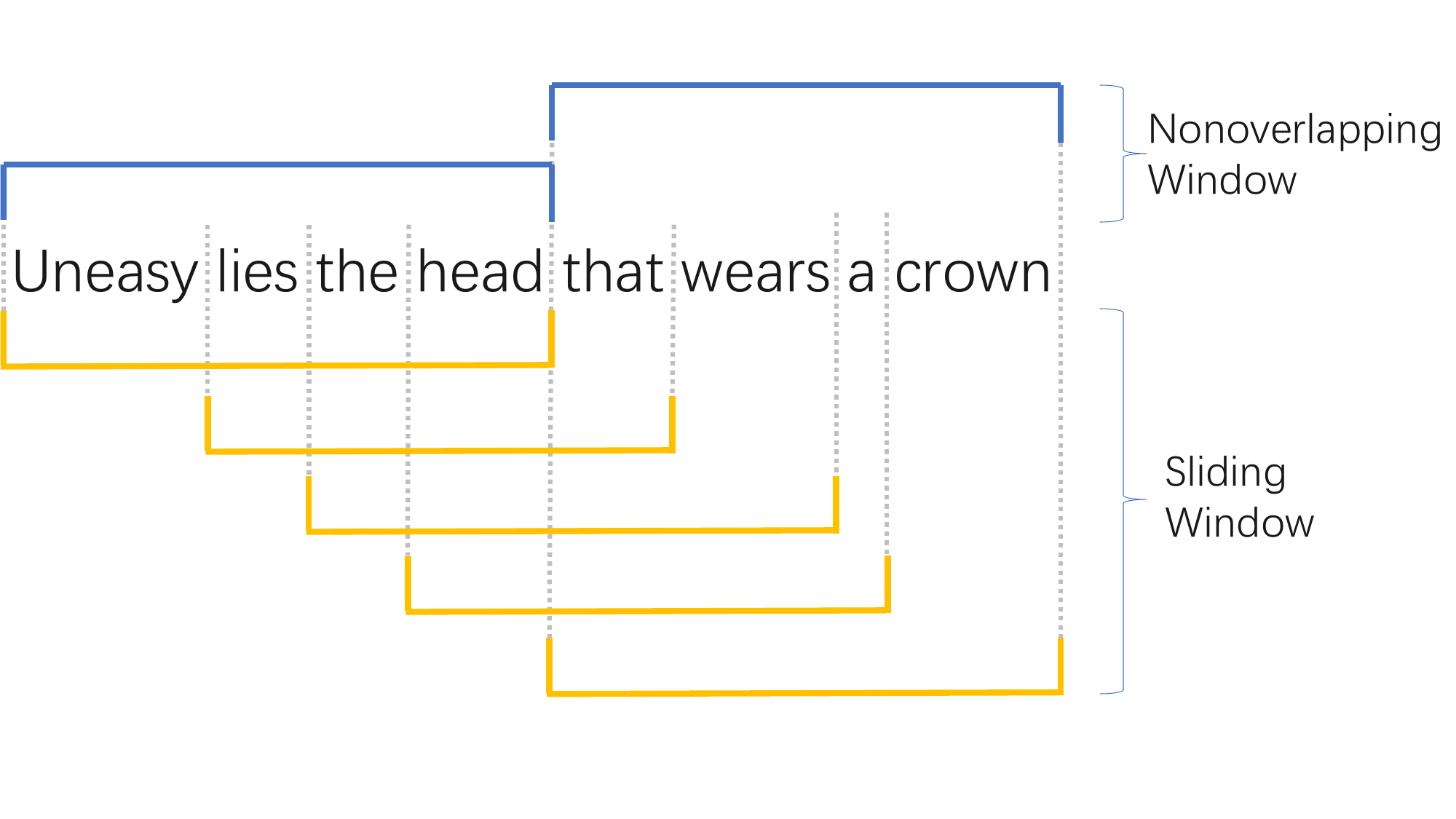}
\vspace{-3mm}
        \caption{\textbf{Sliding window inference \emph{vs} Nonoverlapping inference.} We illustrate the difference between sliding window inference and nonoverlapping inference in the right figure. The left figure shows the curves of ''Sliding Window`` and ''Nonoverlapping Window``corresponding to the ppls calculated by a language model at different inference window sizes. As a reference, we also plot the average ppl of Alibi extrapolated from 512 to 9216. The sliding window inference has much lower ppl than the nonoverlapping one and is close to the Alibi's performance.}
        \vspace{-3mm}
        \label{fig:window}
\end{figure*}

\begin{table}
    \centering
    \small
     \tabcolsep=0.65cm
    \caption{\textbf{Relative average inference time.} We compute the relative average inference time of sliding window inference and nonoverlapping inference over a set of window sizes \{16,32,64,128,258,512\}. We also include the Alibi inference time as a reference.}
    \vspace{-3mm}
    \label{table:speed}
    \begin{tabular}{l|c}
    \hline
        Method & Rel Avg infer time \\ \hline
        Sliding Window & 44.35 \\ 
        Nonoverlapping Window & 1.00 \\ 
        Alibi & 1.00 \\ \hline
    \end{tabular}
\end{table}

There are two ways to infer window attention: nonoverlapping inference and sliding window inference as shown on the right of Figure~\ref{fig:window}. In sliding window inference, the tokens within each sliding window must be re-encoded multiple times, making it substantially slower than the nonoverlapping one. In table~\ref{table:speed} we compare the average inference time over a group of window sizes between the sliding window inference and nonoverlapping window inference. The sliding window one is more than 44 times slower than the nonoverlapping one. However, as shown on the left of Figure~\ref{fig:window}, the sliding window inference has much lower ppl than the nonoverlapping one. 

\section{Transformer Extrapolation Exploration}
In this section, we first describe the hypothesis about why existing RPE-based length extrapolation methods can extrapolate sequences in inference and provide empirical evidence for it. Then we derive the conditions for length extrapolation in detail and demonstrate that recent RPE-based length extrapolation methods~\cite{chi2022kerple,alibi} satisfy the conditions.

\begin{figure}
\centering
\small
 \tabcolsep=0.03cm
 \vspace{-3mm}
\begin{tabular}{cc}
\includegraphics[width=0.48\linewidth]{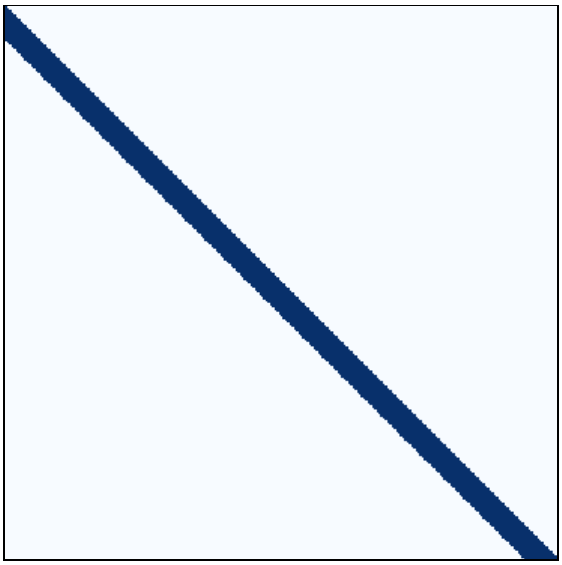} & 
\includegraphics[width=0.48\linewidth] {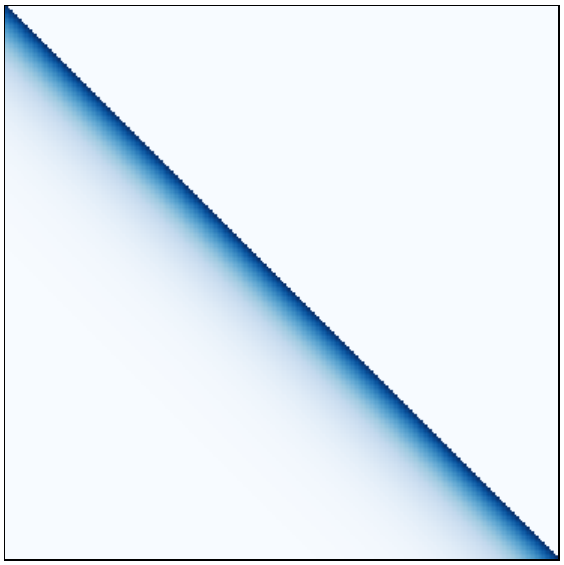} \\
(a) Sliding window & (b)Sliding Window
\end{tabular}
\vspace{-3mm}
        \caption{\textbf{Visualization of attention reweighting.} We plot the reweighting schema of sliding window attention and Alibi linear decay bias. They share a similar behavior in that only neighboring tokens can influence the attention results.}
        \vspace{-1mm}
        \label{fig:attention_map}
\end{figure}

\subsection{The hypothesis}
A sliding window attention with window size $w$ is equivalent to the following RPE on full attention:
\begin{equation}
\mathrm m_{ij}= \begin{cases}
0, & i-j \le w .\\
-\infty, & \mathrm{others}.
\end{cases}
\label{swarpe}
\end{equation}
By comparing Eq.~\ref{swarpe} and the corresponding RPE of Alibi~\cite{alibi} in Figure~\ref{fig:attention_map}, we can see that they both have the same behavior in that they both concentrate tokens inside a specified range. Also, in Figure~\ref{fig:window}, we show that the performance of Alibi is similar to the sliding window attention when the window size is sufficiently large. Based on these two observations, we make the following hypothesis:
\newtheorem{main-conjecture}{Hypothesis}[section]
\begin{main-conjecture}
\label{main-conjecture}
A RPE that makes a transformer extrapolatable needs to have similar behavior to sliding window attention, \ie $\delta(i, j)$ should satisfy:
\begin{equation}
    \forall \epsilon > 0, \exists j_0,  s.t, j> j_0, 
 \delta(i, j) < \epsilon,
\label{conjecture-eq}
\end{equation}
where $\delta(i, j)\triangleq \| \mathbf o_i^i - \mathbf o_i^j\|$, and the window length $j$ needs to be sufficiently large.
\end{main-conjecture}
In the following sections, we will derive the conditions for RPEs that satisfy \eq~\ref{conjecture-eq}.

\subsection{The conditions}
Let us introduce the first lemma:
\newtheorem{model-extrap}[main-conjecture]{Lemma}
\begin{model-extrap}
\label{model-extrap}
When the following condition is satisfied, \eq~\ref{conjecture-eq} holds.
\small  
\begin{equation}
\label{model_cond}
{\lim}_{i\to \infty}C_{ii} \triangleq C <\infty.
\end{equation}
\normalsize
\end{model-extrap}
\begin{proof}[Proof]
When $ i\le m$, 
the test sequence length is smaller than the max sequence length $m$ during training, take $j=i$, we get
$\| \mathbf o_i^{i} - \mathbf o_i^j\|=\| \mathbf o_i^{i} - \mathbf o_i^i\|=0.
$
When $i>m$, we can reformulate \eq~\ref{att_trun} as:
\small 
\begin{equation*}
\begin{aligned}
\mathbf o_i^{i}
&=\frac{\sum_{i-j+1\le s\le i}c_{is} \mathbf  v_s + \sum_{1\le s\le i-j}c_{is}\mathbf  v_s}
{C_{ii} }\\
&=\frac{\sum_{i-j+1\le s\le i}c_{is} \mathbf  v_s }{C_{ij} }
\frac{{C_{ij} }}{{C_{ii}}}
+
\frac{ \sum_{1\le s\le i-j}c_{is}\mathbf  v_s}{C_{ii} - C_{ij}}
\frac{C_{ii} - C_{ij}}{C_{ii}}\\
&= \frac{ \sum_{i-j+1\le s\le i}c_{is}\mathbf  v_s}{ C_{ij}}\frac{{C_{ij} }}{{C_{ii}}} + 
\frac{ \sum_{1\le s\le i-j}c_{is}\mathbf  v_s}{C_{ii} -C_{ij}} \left (
1-\frac{{C_{ij} }}{{C_{ii}}}
\right).
\end{aligned}
\end{equation*}
\normalsize

Therefore we have:
\small
\begin{equation}
\label{delta-eq}
\begin{aligned}
\mathbf o_i^{i} -  \mathbf o_i^j
&=
\left( 1- \frac{{C_{ij} }}{{C_{ii}}} \right)\left(
\frac{ \sum_{i-j+1\le s\le i}c_{is}\mathbf  v_s}{ C_{ij}}- \frac{ \sum_{1\le s\le i-j} c _{is}\mathbf  v_s}{C_{ii} - C_{ij}}
\right).
\end{aligned}
\end{equation}
\normalsize
For the second part:
\begin{equation}
\begin{aligned}
&\left\|
\frac{ \sum_{i-j+1\le s\le i}c_{is}\mathbf  v_s}{ C_{ij}}- \frac{ \sum_{1\le s\le i-j} c _{is}\mathbf  v_s}{C_{ii} - C_{ij}}
\right\|\\
&\le \frac{ \sum_{i-j+1\le s\le i }c_{is} \| \mathbf  v_s\|}{C_{ij}} 
+\frac{ \sum_{1\le s\le i-j} c _{is}\|\mathbf  v_s  \|}{C_{ii} - C_{ij}} \\
&\le \frac{ \sum_{i-j+1\le s\le i }c_{is}l }{ C_{ij}} 
+\frac{ \sum_{1\le s\le i-j} c _{is} l}{C_{ii} - C_{ij}} = 2l
\end{aligned}
\end{equation}
We have
\begin{equation}
\label{delta-bound}
\delta(i,j) \le 2\left(1- \frac{{C_{ij} }}{{C_{ii}}}\right) l.
\end{equation}
According to Eq~\ref{model_cond} and the tail of convergence series is arbitrarily small. $\forall C/2>\epsilon > 0$, we can find a $j_0$, s.t. if $i\ge j > j_0$, $C_{ii}-C_{ij}<\epsilon$.
We can also find a $j_1$, s.t. if $i\ge j > j_1$, $C-\epsilon< C_{ii} < C+ \epsilon$.
If we take $j_2=\max(j_0, j_1)$, so if $i\ge j \ge j_2$, we have:
\begin{equation}
    C_{ii}-C_{ij}<\epsilon,
C-\epsilon < C_{ii} < C+\epsilon
\end{equation}
So when $i\ge j \ge j_2$, we have
\begin{equation}
\begin{aligned}
\delta(i,j) \le 2\left(1- \frac{{C_{ij} }}{{C_{ii}}}\right) l
&= 2 \frac{C_{ii}-C_{ij}}{C_{ii}} l
\le 2 \frac{\epsilon}{C-\epsilon}l\\
&\le \frac{2l\epsilon}{C-C/2}=\frac{4l\epsilon}{C}
\end{aligned}
\end{equation}
According to the definition of limitation, \eq~\ref{model_cond} holds.
\end{proof}

This lemma implies that for any token if the attention of the model focuses on its neighboring $j(j\ge j_2)$ tokens, the model has length extrapolation property. 
The lemma accompanies our intuitions. Does it mean that {as long as} a RPE follows the same principle, \ie places more weights on neighboring $j$ tokens, the model is guaranteed to have the length extrapolation property?
In the following sections, we will demonstrate that concentrating more weights on neighboring tokens does not guarantee the transformer has the length extrapolation property.
Specifically, we will provide a mathematical proof of the sufficient conditions for RPE to have the length extrapolation property.

\newtheorem{rpe-conv}[main-conjecture]{Theorem}
\begin{rpe-conv}
\label{rpe-conv}
When the following condition is satisfied, \eq~\ref{conjecture-eq} holds.
\small
\begin{equation}
\label{rpe_cond}
{\lim}_{i\to \infty}B_{ii} < \infty, 
B_{ii}= \sum_{1\le t\le i} b _{it}<\infty.
\end{equation}
 \normalsize
\begin{proof}[Proof]
Since we assume $\|\mathbf q_i\|\le l, \|\mathbf k_i \|\le l$, then:
\small
\begin{equation}
\label{bound}
\begin{aligned}
&a_{ij}= \exp(\mathbf q_i^{\top} \mathbf k_j) \le \exp(l^2),
\end{aligned}
\end{equation}
\begin{equation}
\begin{aligned}
   c_{ij}= a_{ij}  b_{ij}\le \exp(l^2) b_{ij} ,C_{ii}
\le \exp(l^2)  B_{ii}.
\end{aligned}
\end{equation}
 \normalsize
Therefore, \eq~\ref{model_cond}\ can be derived from \eq~\ref{rpe_cond}. Combine with Lemma~\ref{model-extrap}, the proof is concluded.
\end{proof}
\end{rpe-conv}

\newtheorem{rpe-extrap-simp}[main-conjecture]{Theorem}
By leveraging the property of RPE, Theorem~\ref{rpe-conv} can be further simplified as:
\begin{rpe-extrap-simp}
\label{rpe-extrap-simp}
When the following condition is satisfied, \eq~\ref{conjecture-eq} holds.
\small
\begin{equation}
\label{rpe-conv-eq}
\lim_{i\to\infty} \sum_{t=1}^i b_{i-t}=\lim_{i\to\infty} \sum_{t=0}^{i-1}b_t < \infty.
\end{equation}
 \normalsize
\begin{proof}[Proof]
According to the definition of RPE:
\small
\begin{equation}
  B_{ii}= \sum_{1\le t\le i} b _{it}
    =\sum_{t=1}^i b_{i-t}=\sum_{t=0}^{i-1} b_{t}.
\end{equation}
 \normalsize
This means that \eq~\ref{rpe_cond} is equivalent to:
\small
\begin{equation}
{\lim_{i\to \infty}} B_{ii}
={\lim_{i\to \infty}}\sum_{t=0}^{i-1} b_{t}
< \infty. \qedhere
\end{equation}
 \normalsize

\end{proof}
\end{rpe-extrap-simp}
Theorem~\ref{rpe-extrap-simp} indicates that as long as the series of $\exp(\text{RPE})$ converges, the model is guaranteed to have length extrapolation property. Based on this principle, we can mathematically determine whether an RPE allows for length extrapolation before conducting experiments or designing a variety of RPEs that can do length extrapolation. In  Appendix, we show that previous methods such as Alibi~\cite{alibi}, Kerple~\cite{chi2022kerple}, and Sandwich~\cite{chi2022sandwitch} satisfy our derived conditions for length extrapolation.

\subsection{Theoretical receptive field}
\vspace{-1mm}
In the previous section, we established the conditions for length extrapolation. As an extra bonus, we can derive Theoretical Receptive Fields (TRF) for any RPE-based length extrapolation method. Let us start with the definition of Empirical Receptive Field (ERF). ERF can be viewed as a window containing the vast majority of the information contained within the attention. 

Recall \eq~\ref{delta-bound}, 
by setting $1- \frac{{C_{ij} }}{{C_{ii}}} =\epsilon$, we can define:
\small
\begin{equation*}
C_{ij}=C_{ii}(1-\epsilon),\
n_{\text{emp}}(\epsilon)=\inf_{j} (C_{ij} > C_{ii}(1-\epsilon)),
\end{equation*}
 \normalsize
$n_{\text{emp}}(\epsilon)$ is the ERF that represents the minimal sequence length required to maintain the performance within a gap of $\epsilon$. Intuitively, ERF can be viewed as the smallest window that contains the majority of the information within an attention. Since it is related to both $a_{ij}$ and $b_{ij}$, it can only be calculated after training. 

Now we define TRF, which allows us to estimate the receptive field without training. To accomplish this, we consider the upper bound of $C_{ij}$. From the definition of $C_{ij}$ and \eq~\ref{bound}, $C_{ij}$ is upper bounded by $B_{ij}$. Therefore, we can define the TRF $n_{\mathrm{the}}^{b}(\epsilon)$ respect to series $b_t$ as:
\begin{equation}   
\label{definitionTRF}
\begin{aligned}
n_{\mathrm{the}}(\epsilon)
&=\inf_{j} (B_{ij} >B (1-\epsilon))\\ 
&=\inf_{j} \left (\sum_{t=0}^{j-1} b_{t}> B(1-\epsilon) \right)\\ &=\inf_{j} \left (\sum_{t\ge j} b_{t}< B\epsilon  \right)
\end{aligned}
\end{equation}
\normalsize 
where $B=\lim_{j\to \infty}\sum_{t=0}^{j-1}b_t$.
We may find it difficult to give the analytical form of the partial sum of the series at times, but we can still compute the TRF numerically or compare the TRFs of different RPEs using the theorem below:

\newtheorem{general-case}[main-conjecture]{Theorem}
\begin{general-case}
\label{general-case}
If the following conditions hold:
\small
\begin{equation}
\small
\label{general}
\begin{aligned}
\frac{\alpha_t}{\alpha} &\le \frac{\beta_t}{\beta},  t\to \infty,\ 
\alpha \triangleq \lim_{j\to \infty}\sum_{t=0}^{j-1} \alpha _t,  \
\beta \triangleq \lim_{j\to \infty}\sum_{t=0}^{j-1} \beta_t  .
\end{aligned}
\end{equation}
 \normalsize
Then:
\small
\begin{equation}
n_{\mathrm{the}}^{\alpha}(\epsilon) \le n_{\mathrm{the}}^{\beta}(\epsilon), \epsilon \to 0.
\end{equation}
 \normalsize
\end{general-case}

\begin{proof}[Proof]

According to \eq \ref{general}, there exists $t_0 >0$, such that, when $t> t_0$, we have:
\small
\begin{equation}
\frac{\alpha_t}{\alpha} \le \frac{\beta_t}{\beta}.
\end{equation}
\normalsize
Let $\epsilon < \epsilon_0$, where
\small
\begin{equation}
n_{\mathrm{the}}^{\beta}(\epsilon_0)= t_0,
\end{equation}
\normalsize
then we get:
\small
\begin{equation}
\sum_{t\ge n_{\mathrm{the}}^{\beta}(\epsilon)} \beta_{t}\le \beta\epsilon,
n_{\mathrm{the}}^{\beta}(\epsilon)>t_0.
\end{equation}
\normalsize
Finally:
\small
\begin{equation*}
\begin{aligned}
\sum_{t\ge n_{\mathrm{the}}^{\beta}(\epsilon)} \alpha_{t}
 \le \sum_{t\ge n_{\mathrm{the}}^{\beta}(\epsilon)}\frac {\alpha \beta_{t}} \beta  
 \le \frac {\alpha\beta \epsilon } \beta 
= \alpha\epsilon.
\end{aligned}
\end{equation*}
\normalsize
According to \eq~\ref{definitionTRF}, we have:
\small
\begin{equation}
n_{\mathrm{the}}^{a}(\epsilon) \le n_{\mathrm{the}}^{b}(\epsilon).
\end{equation}
 \normalsize
The $\exp(\text{RPE})$ series follows the same trend as TRF, the smaller the series, the smaller the TRF.
\end{proof}

We provide several examples of how to compute TRF in the Appendix.

\vspace{-2mm}
\subsection{Two new RPEs}
\vspace{-1mm}
\label{RPE-example}
Based on the proven conditions of length extrapolation, we can design infinite kinds of RPEs with the length extrapolation property. Here, we propose two new RPEs to empirically prove the conditions and hypothesis, namely:
\begin{equation*}
\begin{aligned}
    \mathrm{Type1}: b_n&=\frac{1}{n^2}=\exp(-2\ln n),\\ \mathrm{Type2}: b_n&=\exp(-\ln^2n);
    \end{aligned}
\end{equation*}

The corresponding TRF of Type 1 is:
\small
\begin{equation}
\textstyle
\begin{aligned}
B_{ij}&=\sum_{i=0}^{j-1}\frac{1}{(i+1)^2}\approx\int_{1}^{j} \frac{1}{x^2} dx=1-\frac{1}{j},
B= 1.\\
n_{\mathrm{the}}(\epsilon)
&=\inf_{j} \left (B_{ij} > B(1-\epsilon)  \right) \\
&=\inf_{j} \left (1-\frac 1 {j}> 1-\epsilon  \right) 
=\Theta \left(\frac 1 \epsilon \right)
\end{aligned}
\end{equation}
\normalsize
For Type 2, it is difficult to provide the analytical form of its TRF. However, we can prove that the TRF of Type 2 is smaller than the TRF of Type 1 using Theorem~\ref{general-case} and the inequality below:
\small
\begin{equation*}
\forall c_1, c_2 > 0,
\frac{\exp(-\ln^2n)} {c_1} < \frac {1/n^2} {c_2}, n\to \infty.
\end{equation*}

\section{Empirical Validation}
\paragraph{Setting}
All models are implemented in Fairseq~\cite{ott2019fairseq} and trained on 8 V100 GPUs. We use the same model architecture and training configuration for all RPE variants to ensure fairness.  
For Wikitext-103~\cite{1609.07843}, since it is a relatively small dataset, we use a 6-layer transformer decoder structure with an embedding size of 512. For other datasets, in particular, we used a 12-layer transformer decoder structure with an embedding size of 768. The evaluation metric is perplexity (PPL) and the max training length during training is 512. The detailed hyper-parameter settings are listed in Appendix. 
\normalsize
\vspace{-2mm}
\paragraph{Dataset}
We conduct experiments on Wikitext-103~\cite{1609.07843}, Books~\cite{Zhu_2015_ICCV}, Github~\cite{pile} and WikiBook~\cite{2202.08005}. Wikitext-103 is a small dataset containing a preprocessed version of the Wikipedia dataset. It is widely used in many NLP papers. Books has a large number of novels, making it a good corpus for long sequence processing. Github consists of a sizable amount of open-source repositories, the majority of which are written in coding languages. WikiBook is a 22-gigabyte corpus of Wikipedia articles and books curated by~\cite{2202.08005}. This corpus is used to validate the performance of various models on large datasets.

\begin{figure}[H]
 \tabcolsep=0.03cm
 \centering
\begin{tabular}{c}
\includegraphics[width=0.9\linewidth]{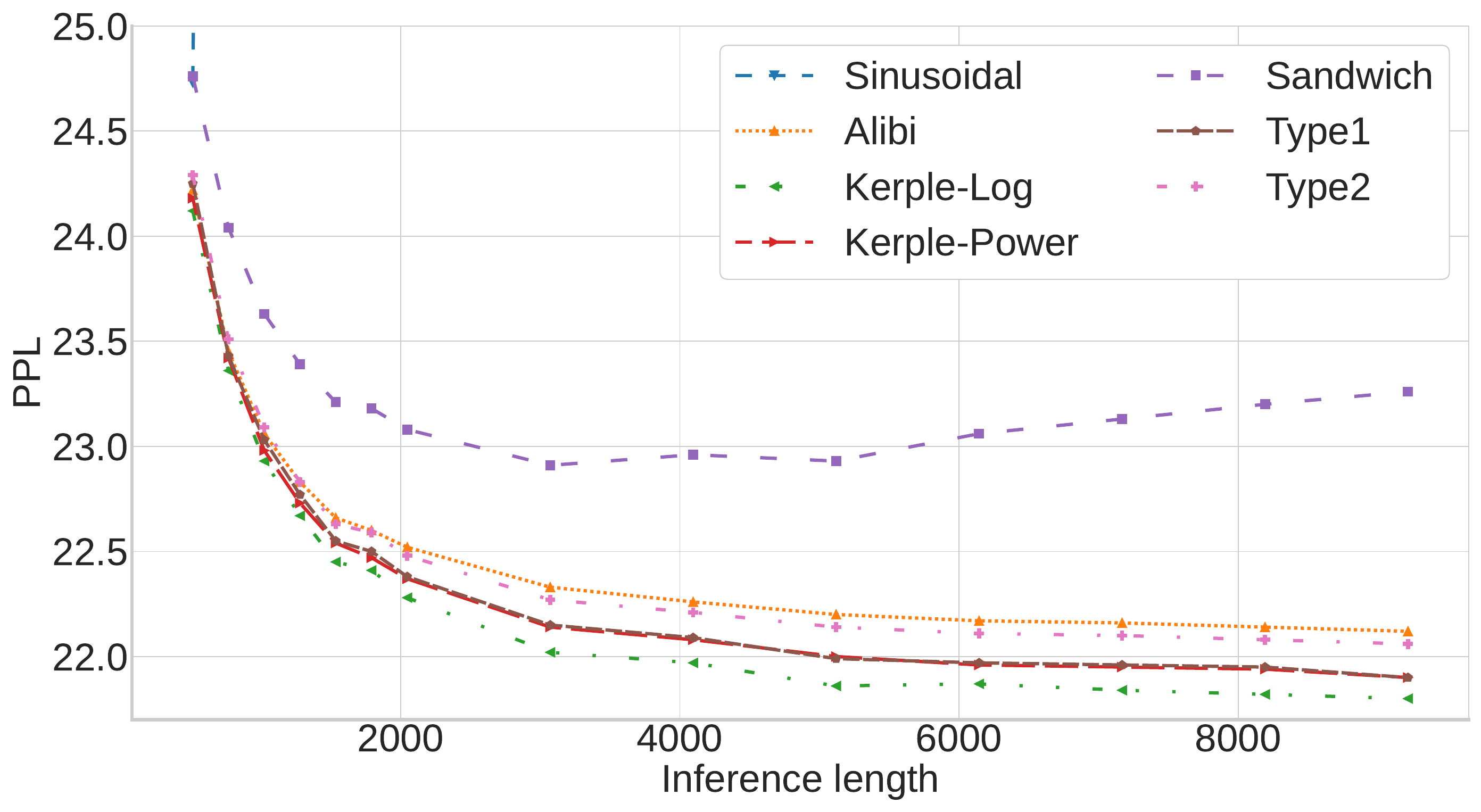} \\
(a) Wikitext-103 \\
\includegraphics[width=0.9\linewidth]{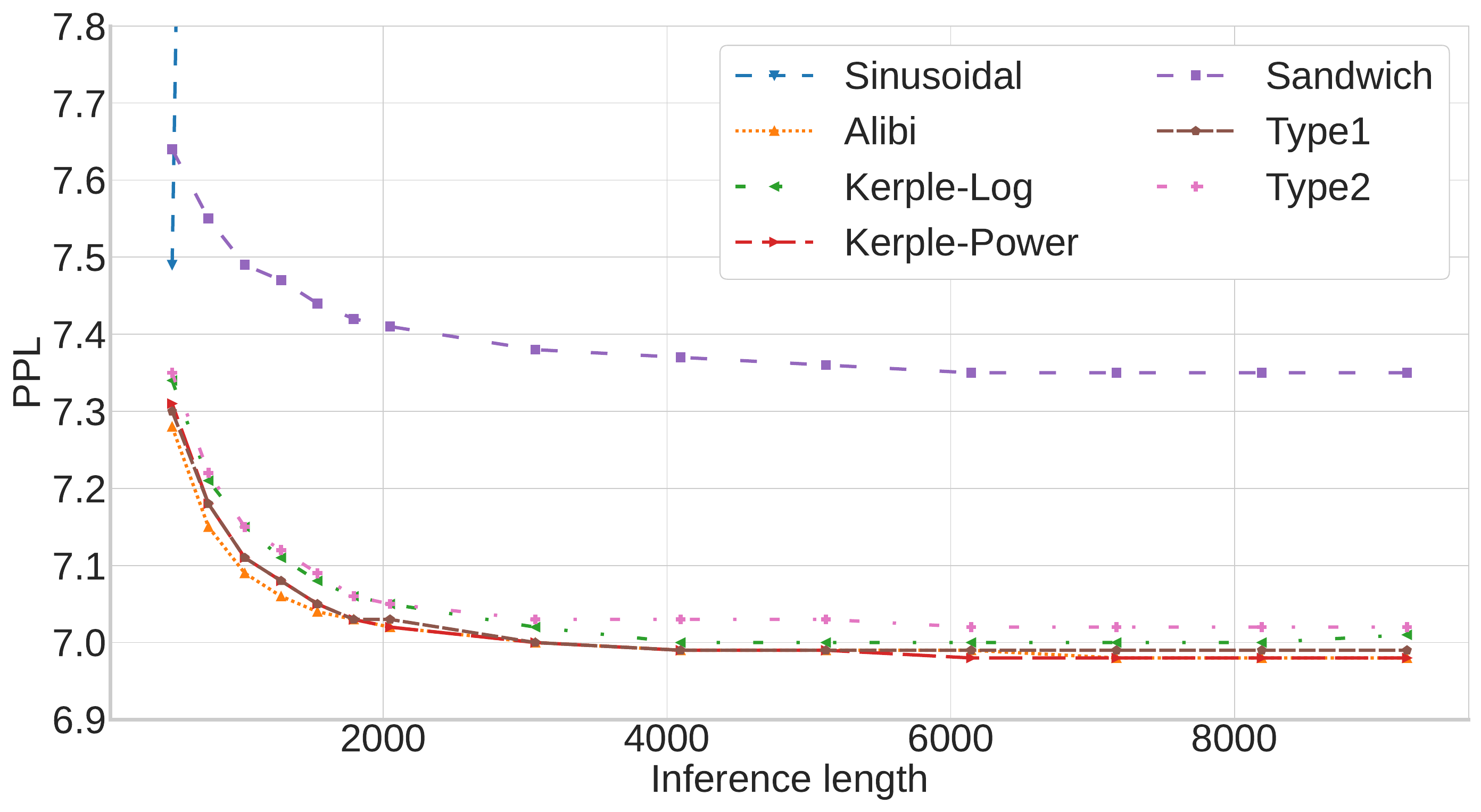} \\
(b) Books \\
\includegraphics[width=0.9\linewidth]{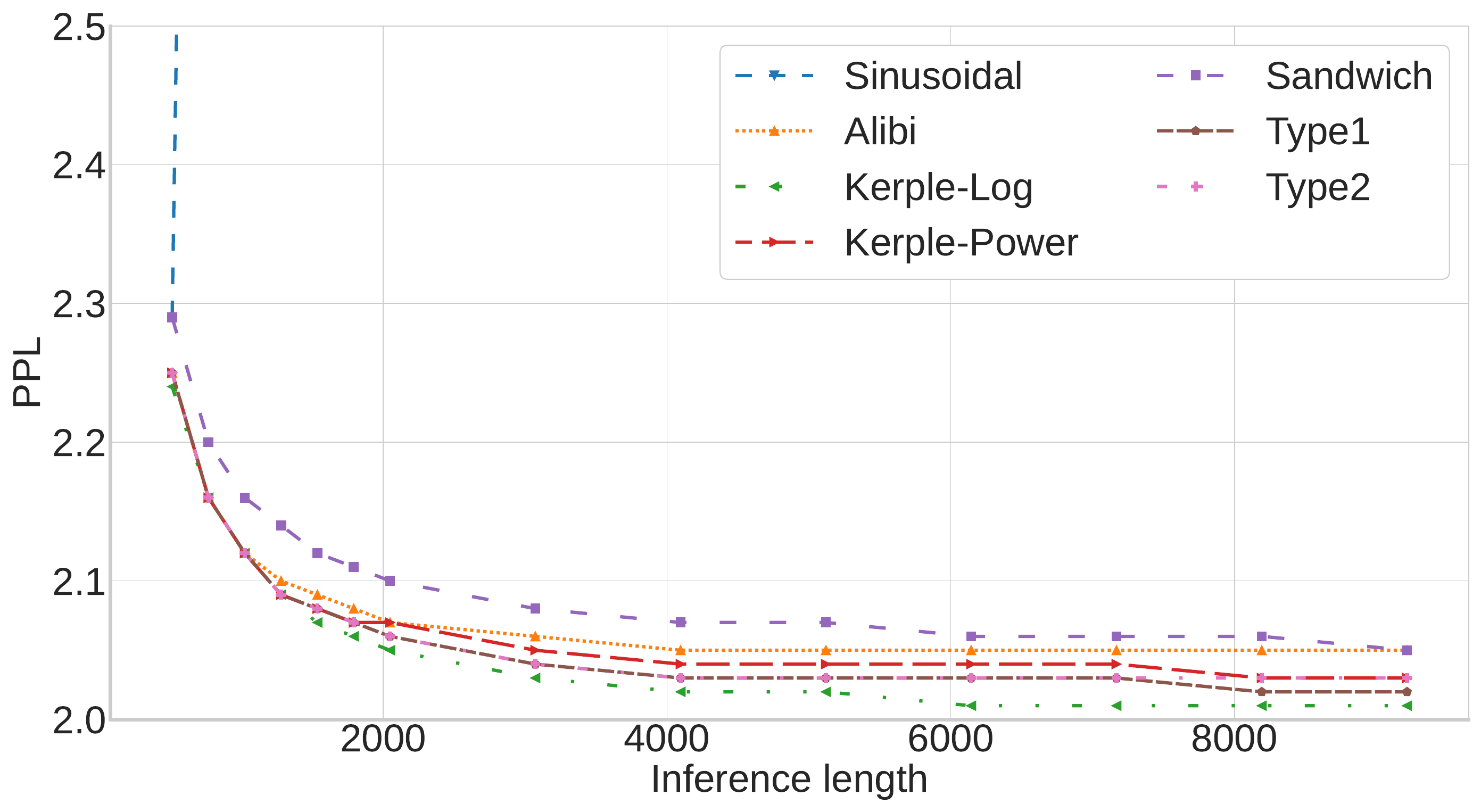} \\
(c) Github \\
\includegraphics[width=0.9\linewidth]{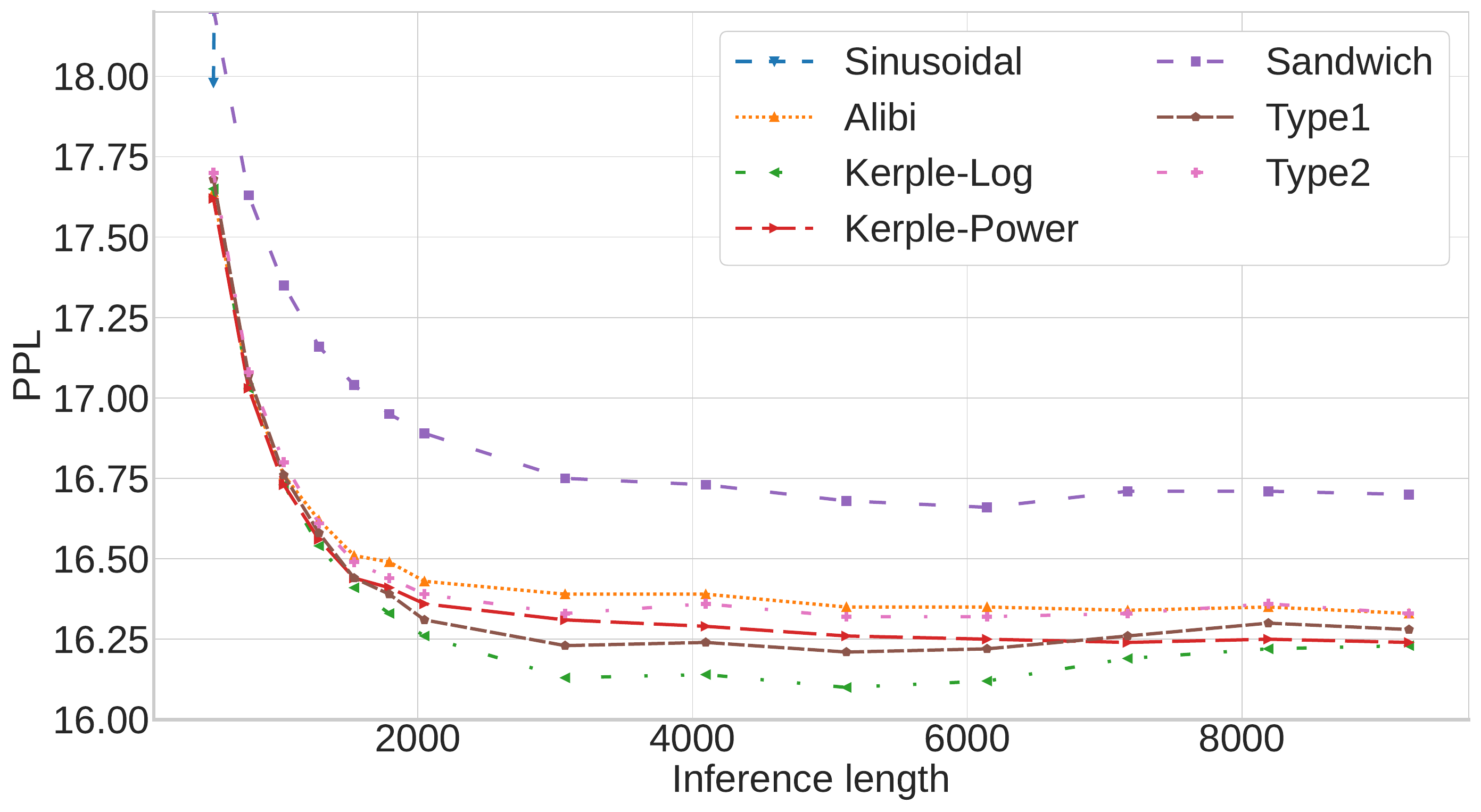} \\
  (d) WikiBook\\
\end{tabular}
\vspace{-3mm}
        \caption{\textbf{Sufficiency validation on Wikitext-103, Books, Github, WikiBook datasets.} To test length extrapolation capability, we lengthen inference sequences from 512 to 9216 tokens and plot the testing PPLs of our proposed Type 1 and Type 2 RPEs, as well as Alibi, Kerple, and Sandwich. All these methods are stable in PPL. For methods that cannot extrapolate, \eg~ Sinusoidal, its PPL grows rapidly.}
        \label{fig: curve}
        \vspace{-3mm}
\end{figure}

\vspace{-2mm}
\paragraph{Validating the sufficiency.}
To empirically validate the sufficiency of our discovered conditions, we integrate the two RPEs that were proposed in the previous section into transformers and test their length extrapolation capability on Wikitext-103, Books, Github, and WikiBook datasets. We
increase the length of the inference sequence from 512 to 9216 tokens and plot the testing PPLs of our proposed RPEs as well as those of existing methods such as Alibi, Kerple, and Sandwich in Figure~\ref{fig: curve}. Detailed numerical results can be found in Table~\ref{table:result1} and Table~\ref{table:result2} from Appendix. All these methods demonstrate good length extrapolation capability. However, the stabilized PPL may vary due to the effectiveness of different positional encoding strategies, which are not considered in this paper. We include the  Sinusoidal~\cite{vaswani2017attention} positional encoding as a reference method that cannot extrapolate, which grows rapidly as the inference sequence length increases.

\begin{figure}[H]
\centering
 \tabcolsep=0.03cm
\begin{tabular}{c}
\includegraphics[width=0.9\linewidth]{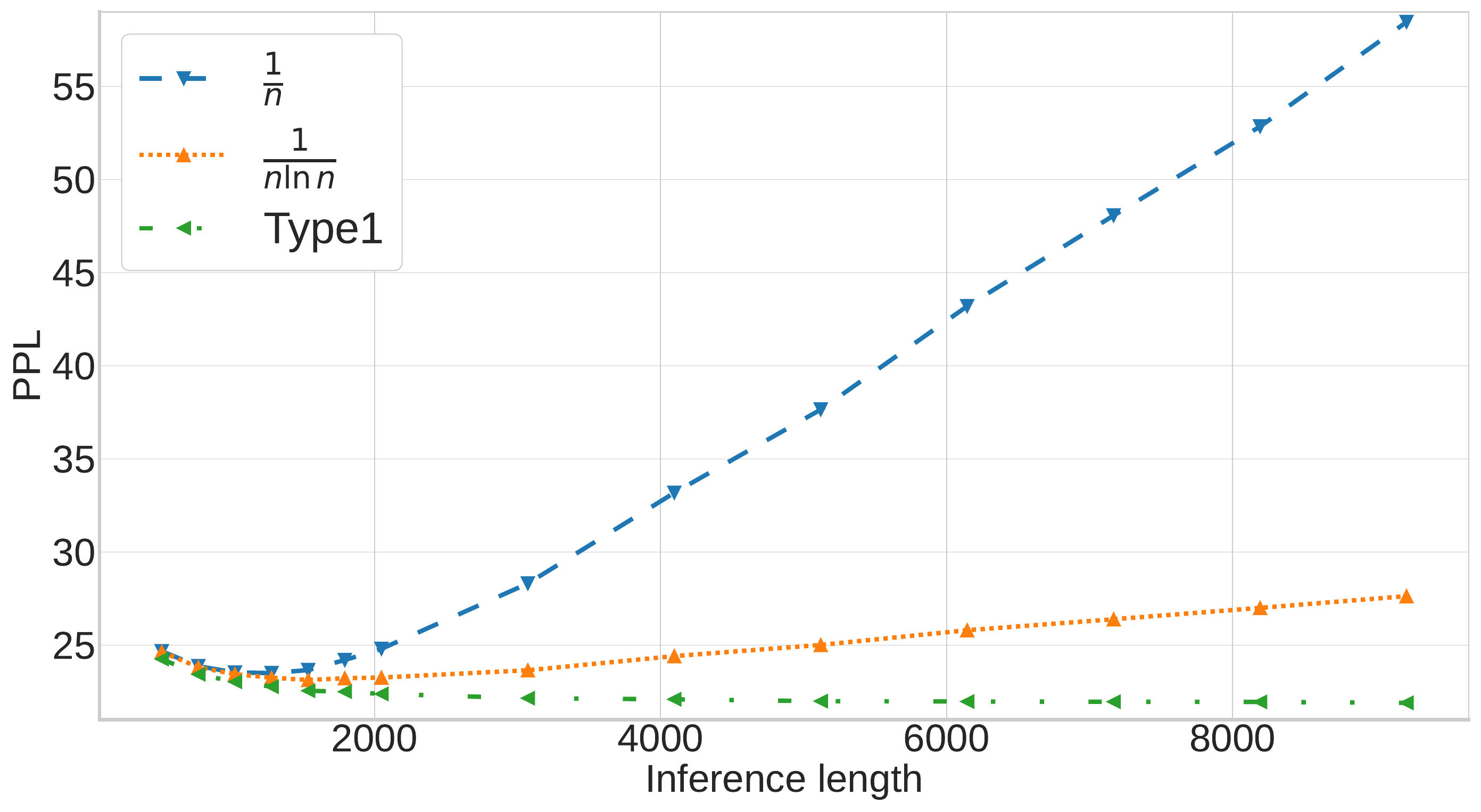}  \\
(a) Wikitext-103 \\
\includegraphics[width=0.9\linewidth]{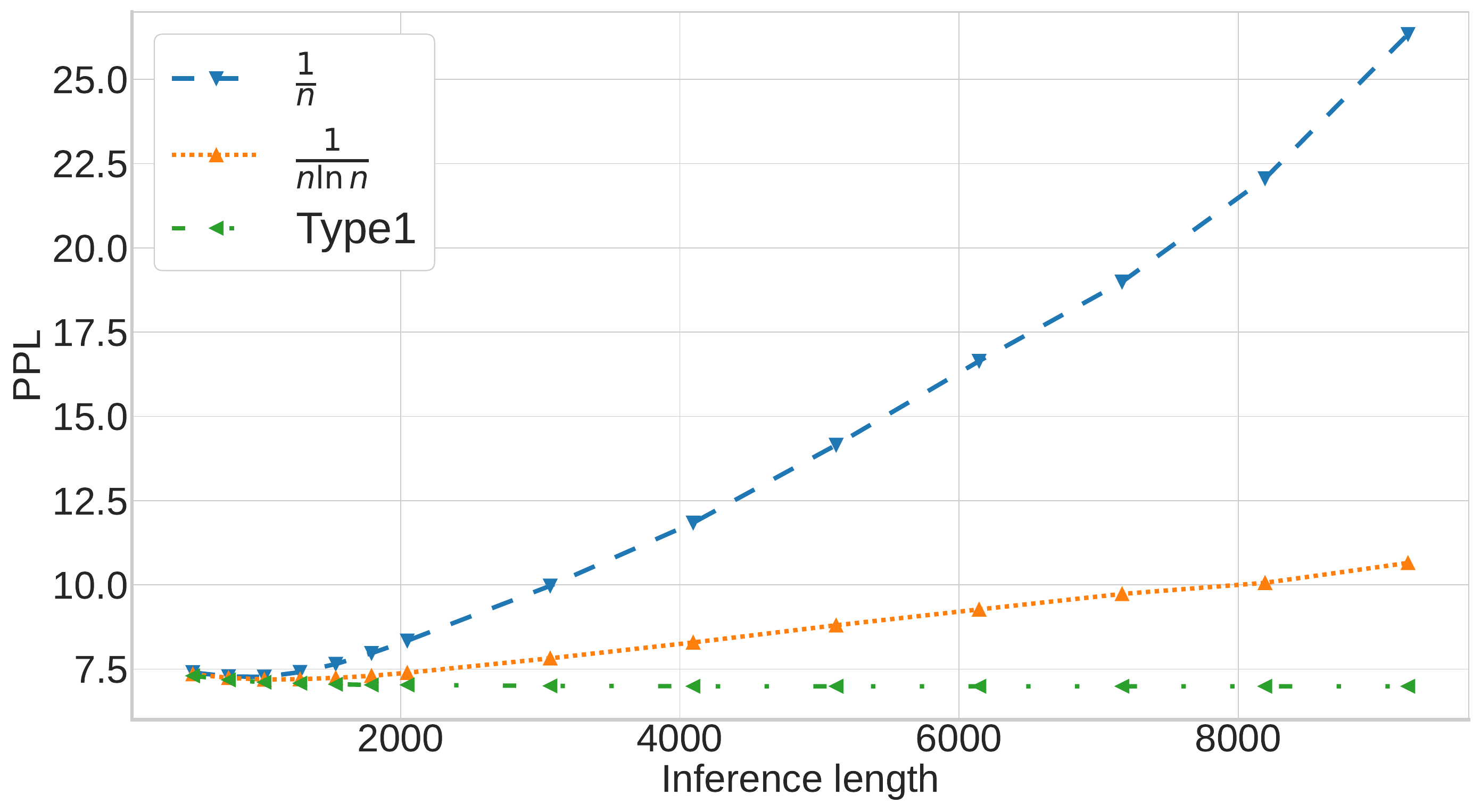} \\
(b) Books \\
\includegraphics[width=0.9\linewidth]{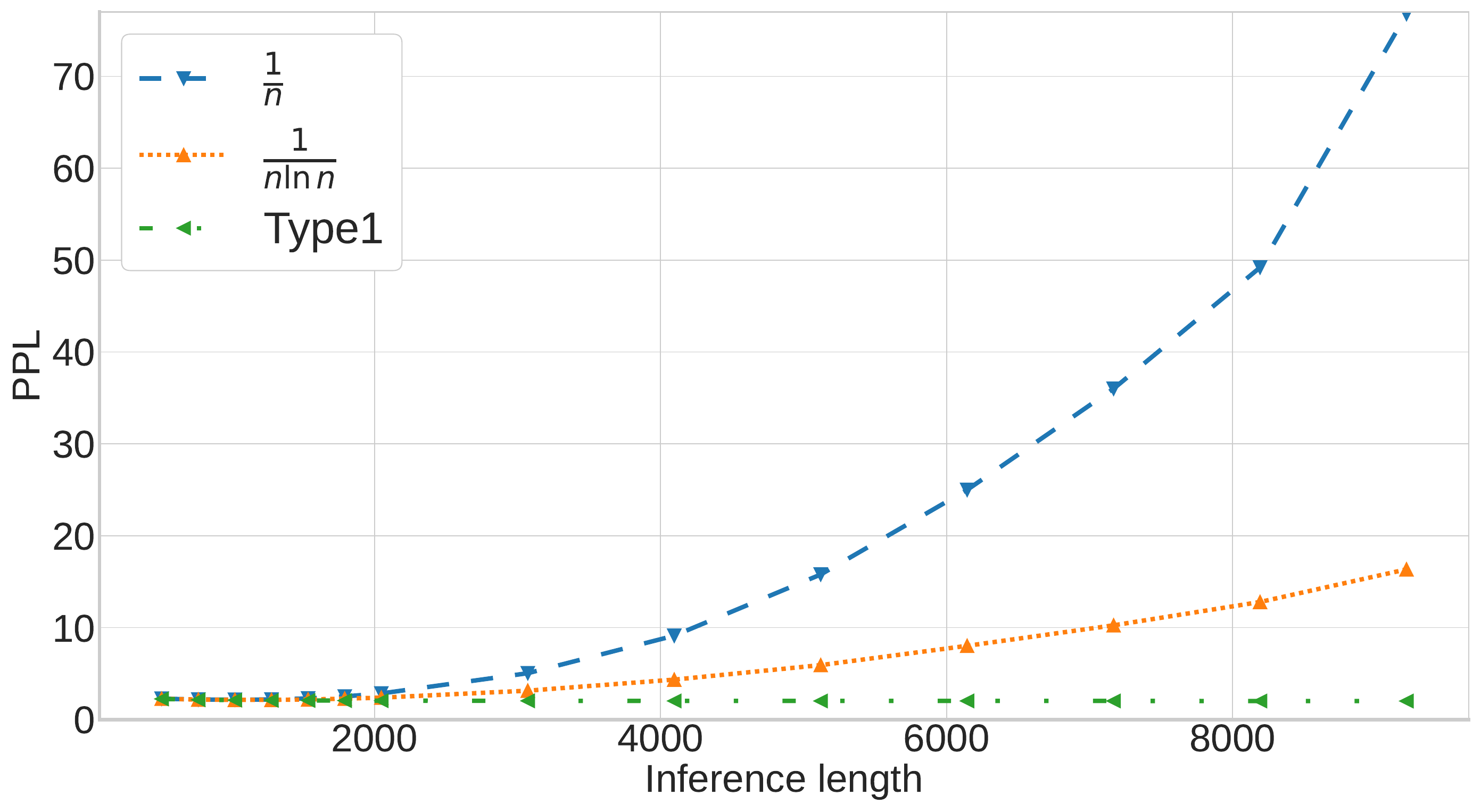}  \\
(c) Github \\
\includegraphics[width=0.9\linewidth]{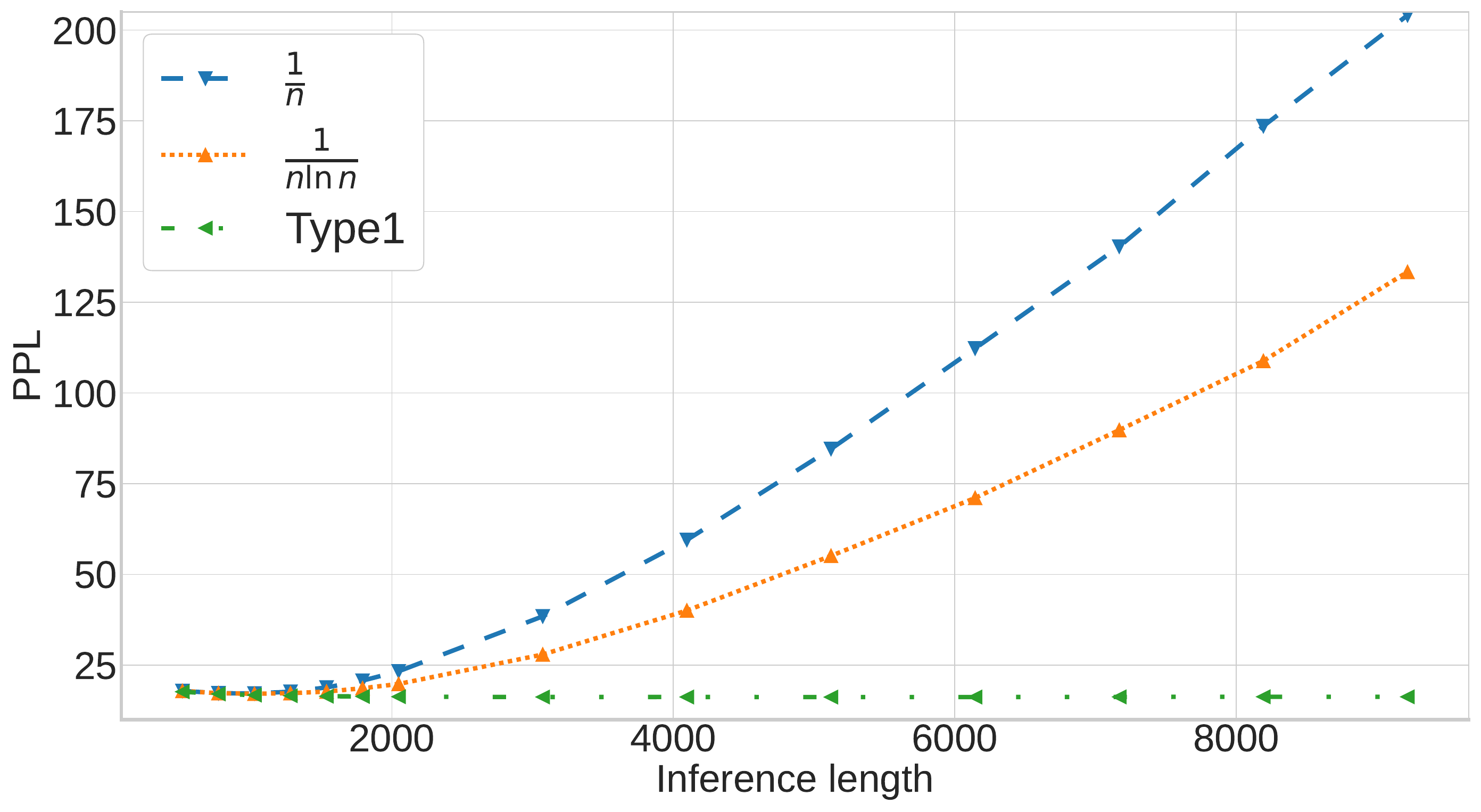} \\ 
  (d) WikiBook\\
\end{tabular}
\vspace{-2mm}
        \caption{\textbf{Necessity validation on Wikitext-103, Books, Github, WikiBook datasets.} We select two RPEs that do not satisfying Theorem~\ref{rpe-extrap-simp}, \eg~$b_n=\frac{1}{n}$ and $b_n=\frac{1}{n\ln n}$. We increase the inference sequence length from 512 to 9216 tokens and plot the testing PPLs for them.
        Their PPLs increase rapidly as the inference sequence lengthens, whereas Type 1 remains stable.
        }
        \vspace{-3mm}
        \label{fig:counter}
\end{figure}

\vspace{-2mm}
\paragraph{Validating the necessity.}
Although we only provide mathematical proof for the sufficiency of our discovered conditions, we also attempt to verify their necessity empirically in this section. Specifically, we pick two RPEs that are very close to satisfying Theorem~\ref{rpe-extrap-simp} as follows. Note that both of them concentrate their weight on neighboring tokens.
\begin{equation*}
\mathrm{Example1}:b_n=\frac{1}{n},\ \mathrm{Example2}:b_n=\frac{1}{n\ln n}
\end{equation*}
Below is a brief mathematical proof that the above RPEs do not satisfy Theorem~\ref{rpe-extrap-simp}.
\small
\begin{equation*}
\begin{aligned}
\sum_{n=1}^{k} \frac 1 n &> \int_{1}^{k+1} \frac 1 x dx=\ln(k+1), \\
\sum_{n=3}^{k} \frac 1 {n\ln n} &> \int_{3}^{k+1} \frac 1 {x\ln x} dx=\ln\ln(k+1) - \ln\ln3.
\end{aligned}
\end{equation*}
\normalsize
We then empirically test their length extrapolation capability on Wikitext-103, Books, Github, and WikiBook datasets by 
\noindent
scaling the inference sequence length from 512 to 9216 tokens. As shown in Figure~\ref{fig:counter}, the PPLs of both RPEs grow rapidly as the length of the testing sequence increases. It de-
\noindent
monstrates that both of them cannot extrapolate. We also include Type~1 RPE in Figure~\ref{fig:counter} as a reference that can extrapolate. Detailed numerical results can be found in Table~\ref{table:result3} 
\begin{figure}[H]
\centering
 \tabcolsep=0.03cm
\begin{tabular}{cc}
\includegraphics[width=0.9\linewidth]{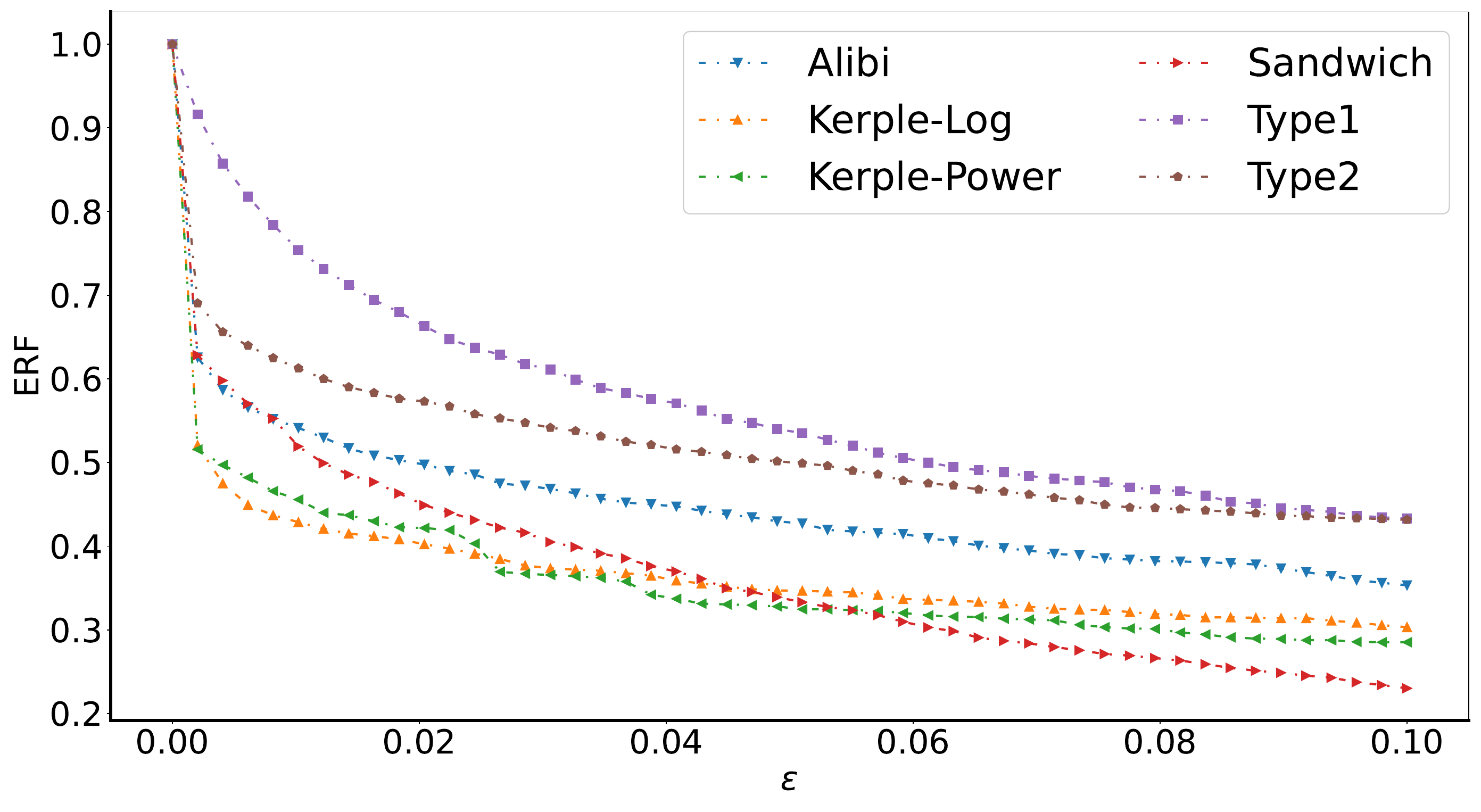}  \\
(a) Wikitext-103 \\
\includegraphics[width=0.9\linewidth]{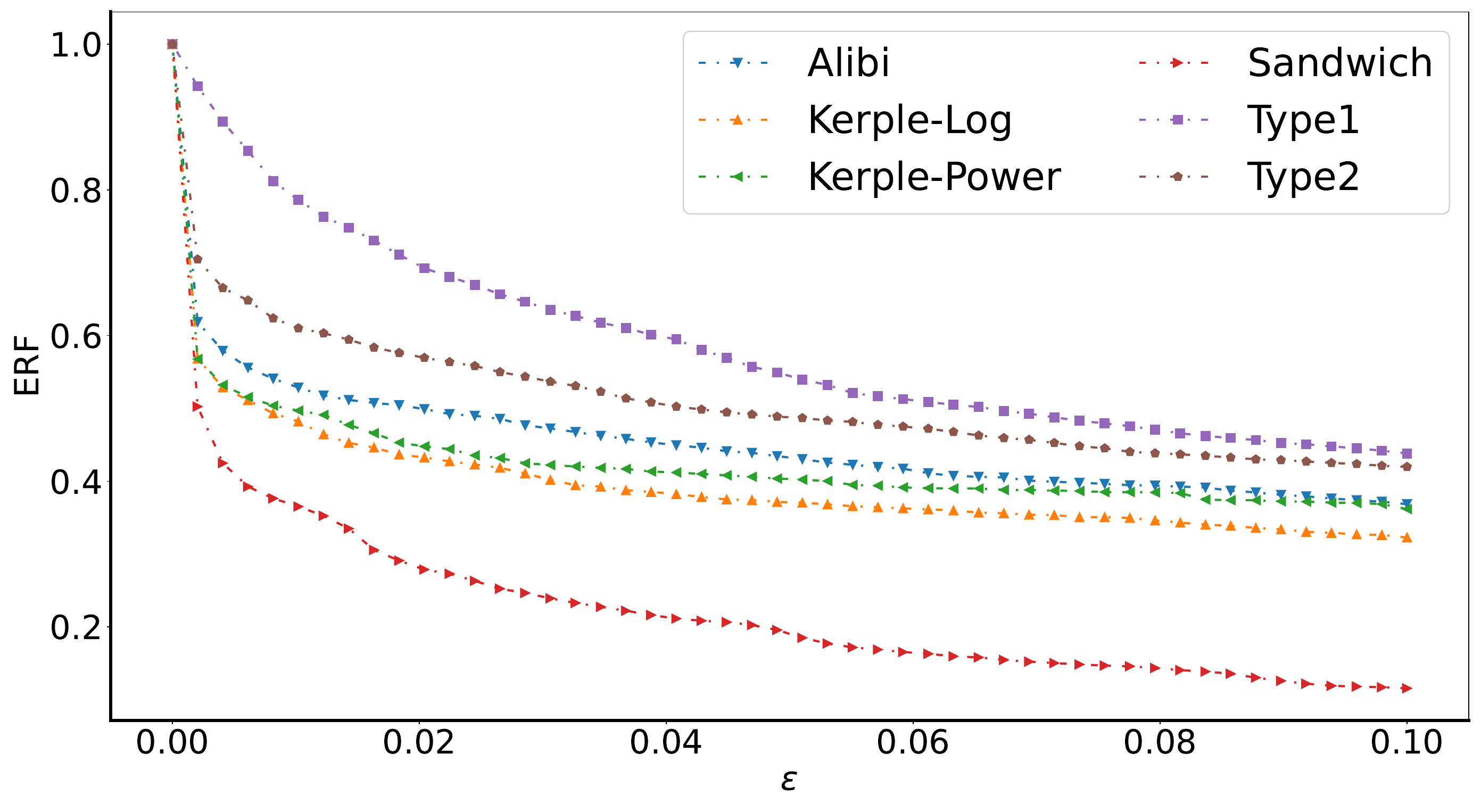} \\
 (b) Books  \\
\includegraphics[width=0.9\linewidth]{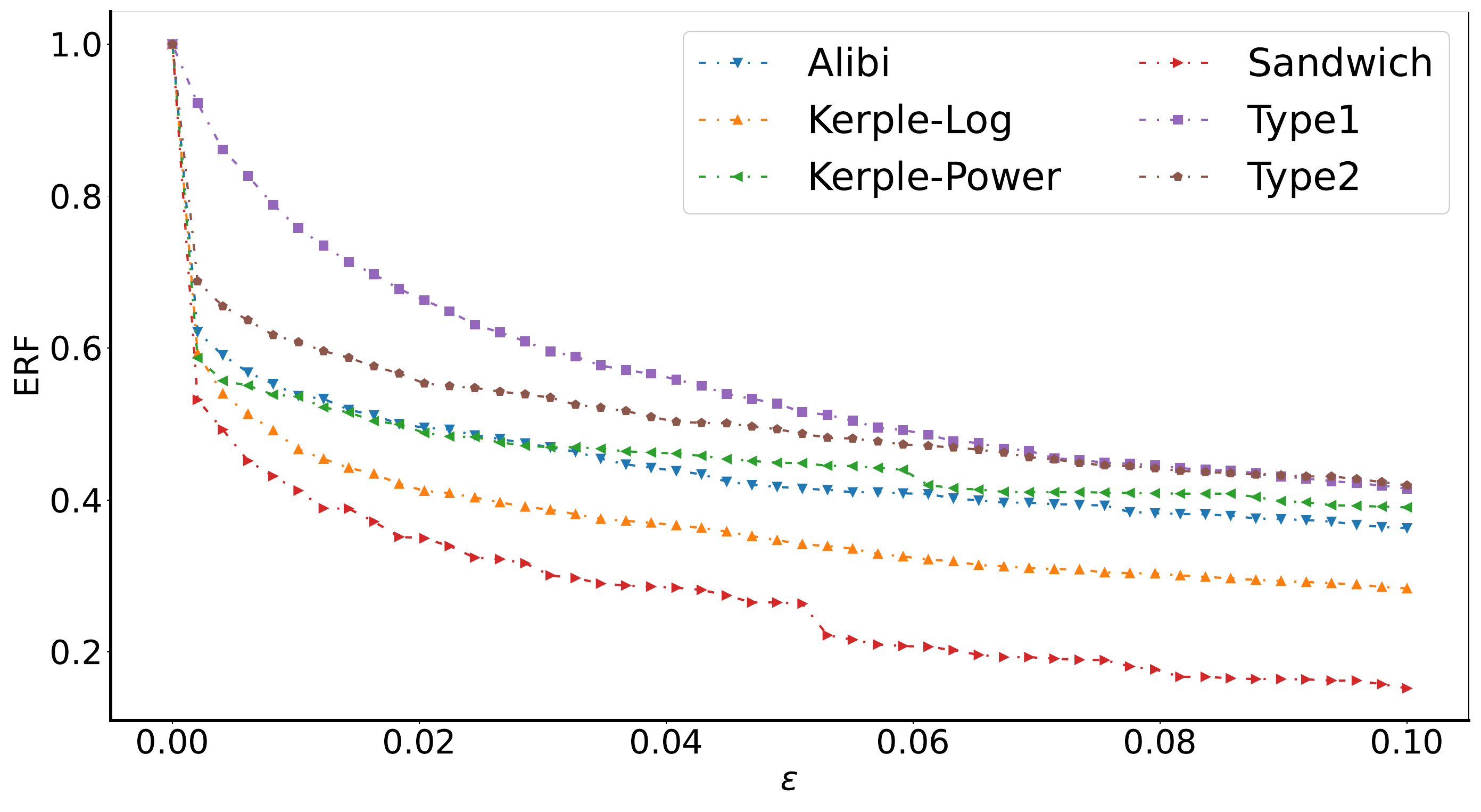} \\
(c) Github \\
\includegraphics[width=0.9\linewidth]{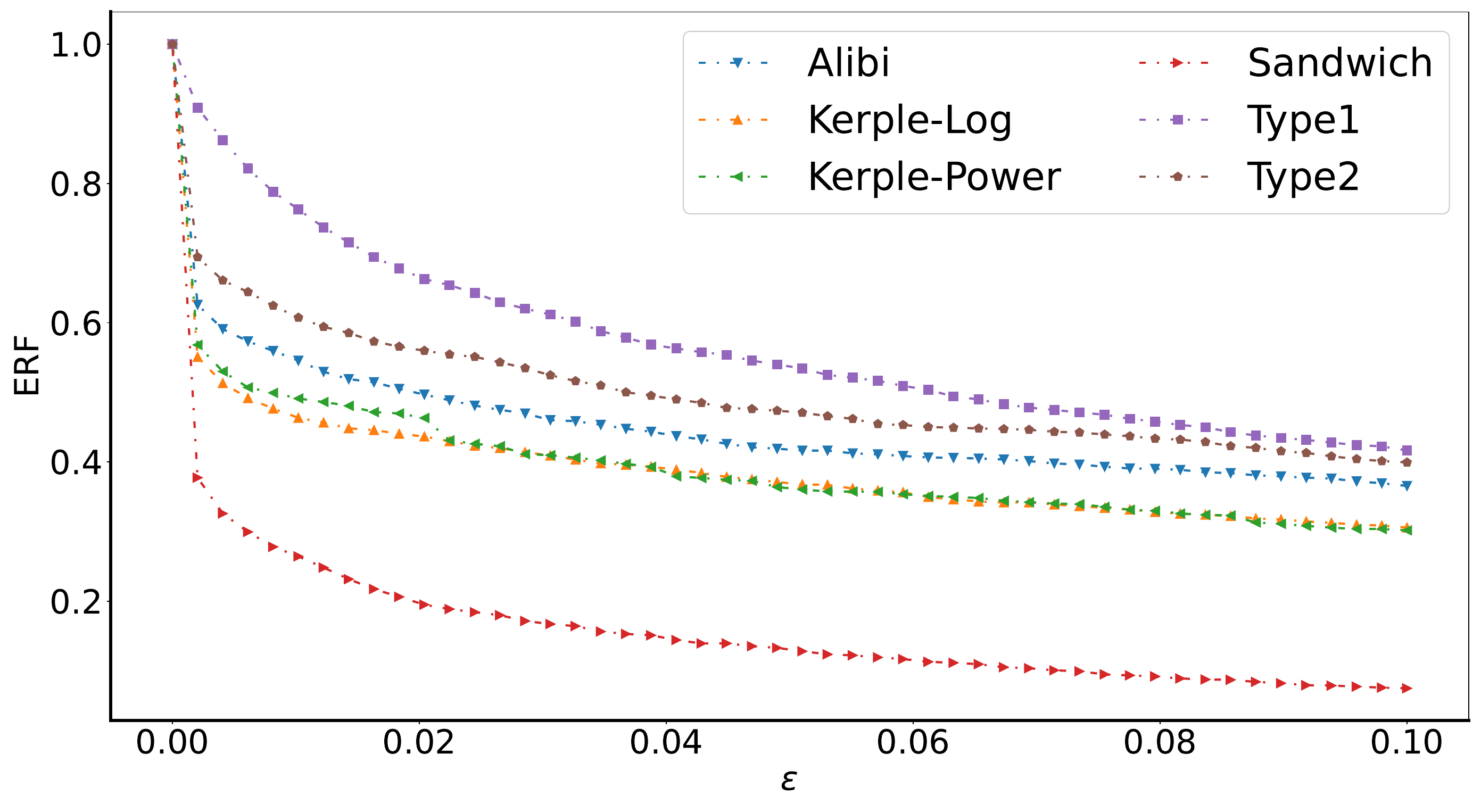}  \\
  (d) wikibook 
\end{tabular}
\vspace{-3mm}
        \caption{\textbf{Visualization of ERF}
        We plot the ERF for Alibi, Kerple, Sandwich and our proposed Type 1 and Type 2 methods on Wikitext-103, Books, Github, and WikiBook datasets using trained models. By comparing with Figure~\ref{fig:trf}, we can find that the overall trend of ERF matches the trend of TRF.
        ERF is normalized for better visualization.}
        \vspace{-3mm}
        \label{fig:erf}
\end{figure}
\normalsize
\noindent
from Appendix.
\begin{figure}[t]
\centering
\includegraphics[width=.9\linewidth]{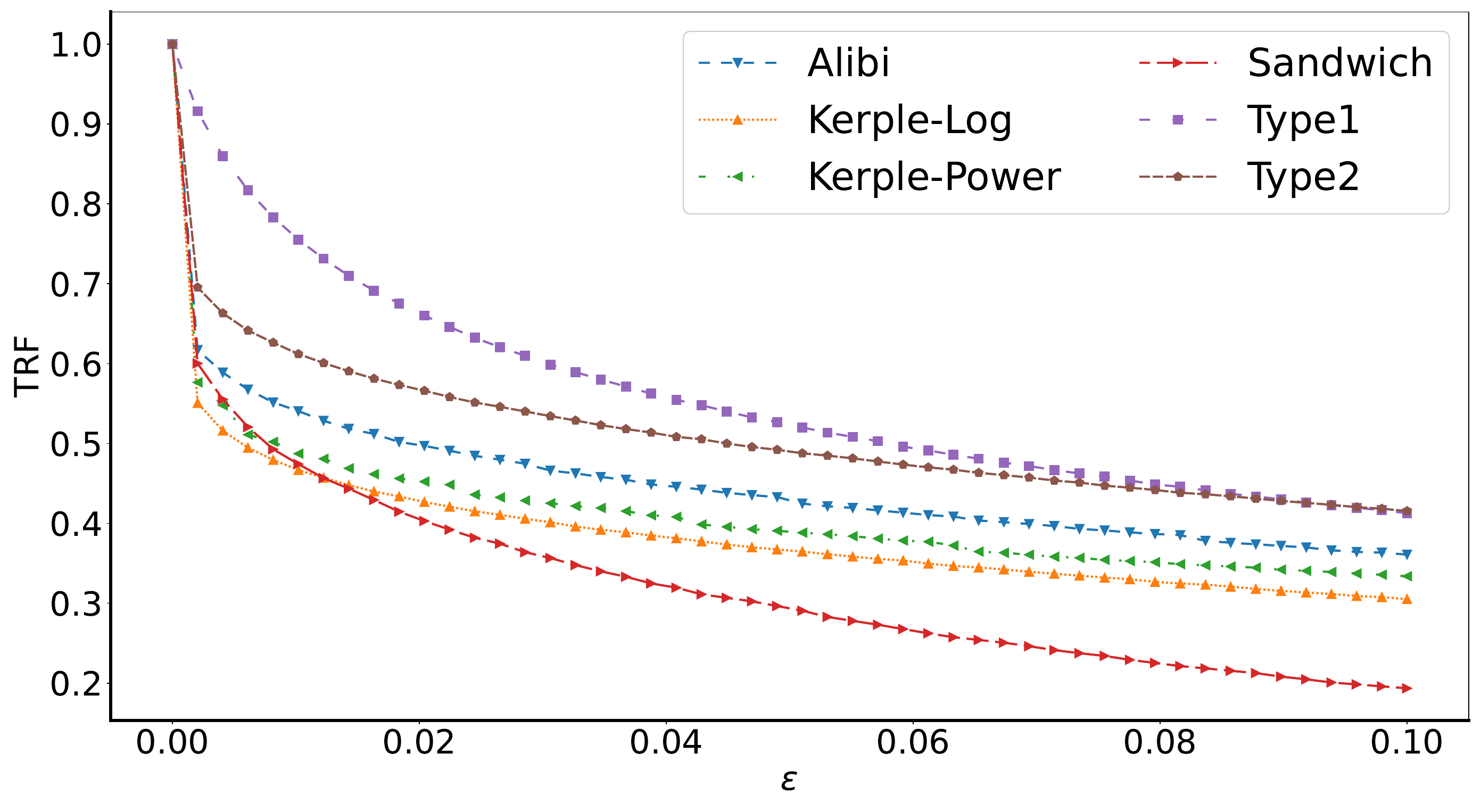} 
\vspace{-3mm}
        \caption{\textbf{Visualization of TRF.}
        We numerically plot TRFs for existing methods and our proposed method. TRF is normalized for visualization. The TRF of Type 1 is larger than Type 2, which matches the Theorem~\ref{general-case} and our analysis.}
        \label{fig:trf}
\end{figure}

\begin{figure}
\centering
 \tabcolsep=0.03cm
\vspace{-2mm}
\begin{tabular}{cc}
\includegraphics[width=0.4\linewidth]{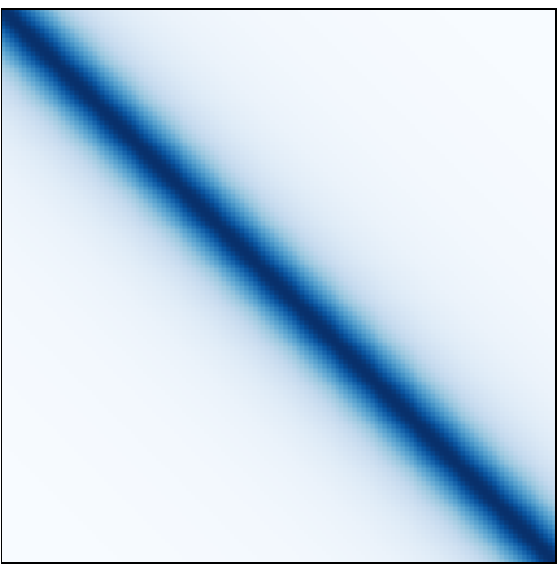} &
\includegraphics[width=0.48\linewidth]{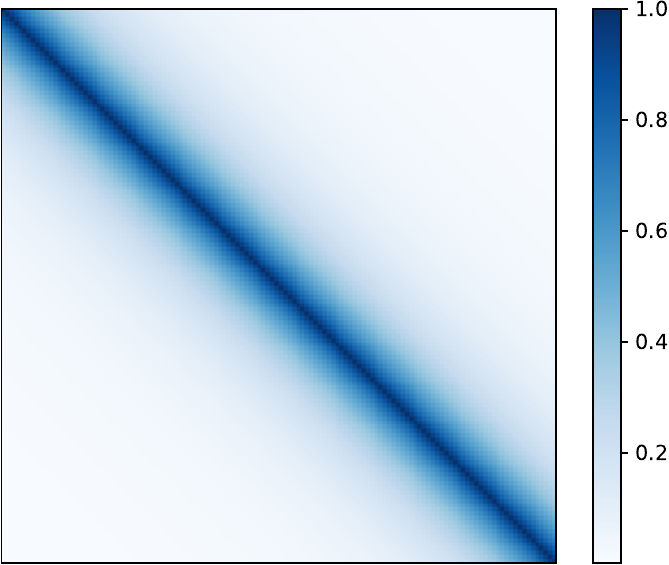} \\
(a) Type1 & (b) Type2
\end{tabular}
\vspace{-2mm}
        \caption{\textbf{Visualization of RPE.} We plot the heatmap of $\exp(\text{RPE})$ for Type 1 and Type 2. Type 2 concentrates weights on closer neighboring tokens than Type 1, indicating a smaller TRF. We also visualize other methods in Appendix.}
        \label{fig:heatmap}
        \vspace{-3mm}
\end{figure}

\vspace{-2mm}
\paragraph{Validating TRF}
We validate our proposed TRF by comparing the trend between the TRF and ERF. We plot the TRFs and ERFs of the Alibi, Kerple, Sandwich, and our proposed RPEs on the aforementioned datasets. As observed in Figure~\ref{fig:trf} and Figure~\ref{fig:erf}, while the curves vary across datasets, TRF estimates a similar overall trend of ERFs.
\vspace{-2mm}
\paragraph{Visualizing RPE}
We visualize the weighting schemes of Type~1 and Type~2 in Figure~\ref{fig:heatmap}, \ie the heatmap of $\exp(\text{RPE})$. 
Type 2 concentrates weights on closer neighboring tokens than Type 1, indicating a smaller TRF and ERF as shown in Figure~\ref{fig:trf} and Figure~\ref{fig:erf}. We also visualize other methods in Appendix.

\section{Conclusion}
In this paper, we explore the secrets of transformer length extrapolation in language modeling. We first make a hypothesis about extrapolation  and then derived the 
sufficient conditions for RPE to have the length extrapolation property. A thorough mathematical analysis reveals that a transformer model is certain to be capable of length extrapolation if the series that corresponds to the exponential of its RPE converges. This observation brings an extra bonus: we can estimate TRFs of RPEs solely based on their formulations. We chose two new RPEs that satisfy the conditions and two that do not to empirically prove the sufficiency of the conditions on four widely used datasets. We also validated our TRFs by comparing them with ERFs on these datasets as well. The results show that our TRFs can accurately reflect the actual receptive fields of RPEs before training. 

\section*{Acknowledgement}
This work is partially supported by the National Key R\&D Program of China (NO.2022ZD0160100).
\bibliography{aaai24}
\input{appendix}

\end{document}

%% file: appendix.tex
\appendix
\onecolumn
\begin{center}
\textbf{\large Supplementary Material}
\end{center}

\section{Mathematical notations}
\label{math notation}
\begin{table}[!ht]
\centering
\small
\setlength{\tabcolsep}{2.1cm}{
\begin{tabular}{c|c}
\hline\hline
\textbf{Notation} & \textbf{Meaning} \\
\hline\hline
$\mathbf{X}$ & Hidden state. \\
$\mathbf{Q},\mathbf{K},\mathbf{V}$ & Query, key, value. \\
$\mathbf{O}$ & Attention output. \\
$d$ & Feature dimension. \\
$\mathbf{m}^\top_s$ & $s$-th row of matrix $M$. \\
\hline\hline
\end{tabular}}
\caption{Mathematical notations used in the paper.}
\end{table}

\section{Examples}
\label{example}
In this section, we use Alibi~\cite{alibi}, Kerple~\cite{chi2022kerple}, and Sandwich~\cite{chi2022sandwitch} as examples to support the discovered sufficient conditions~\ref{rpe_cond} for length extrapolation.

\textbf{Alibi}
The form of Alibi can be written as:
\small
\begin{equation}
b_t =\exp(-k t), k > 0.
\end{equation}
 \normalsize
According to the convergence of geometric series:
\small
\begin{equation}
\lim_{i\to\infty}B_{ii}=\sum_{t=0}^{i-1} \exp(-kt)< \infty,
\end{equation}
 \normalsize
which satisfies our observed conditions.

\textbf{Kerple}
Kerple proposes two forms of RPEs: \emph{log} and \emph{power}. We discuss them separately. 

The formulation of the \emph{Log} variant can be expressed as: 
\small
\begin{equation}
b_t =\exp(-r \log\left(1+ k t\right))=\frac{1}{(1+kt)^{r}}
\end{equation}
 \normalsize
where r,k>0r, k > 0. In Kerple, r,kr, k are learnable. 
Based on Theorem~\ref{rpe-extrap-simp}, to enable the model to extrapolate, we must add the restriction that r>1r > 1 because:
\small
\begin{equation}
\frac{1}{(1+kt)^{r}} \sim \frac 1 {k^{r}t^{r}}, 
\end{equation}
 \normalsize
In empirical analysis, we will show that when r=1r= 1, the model cannot extrapolate. We also checked the trained model from Kerple and found this condition is met.

The \emph{Ploy} variant can be written as:
\small
\begin{equation}
b_t =\exp(-k t^r),0<r\le 2.
\end{equation}
 \normalsize
Since
\small
\begin{equation}
b_t\le \exp(-k t), t\to \infty.
\end{equation}
 \normalsize
according to the convergence of geometric series, we have:
\small
\begin{equation}
\lim_{i\to\infty}B_{ii}=\lim_{i\to\infty}\sum_{t=0}^{i- 1} \exp(-kt^r) < \infty.
\end{equation}
 \normalsize
which satisfies our observed conditions.

\textbf{Sandwich}
Given the formulation of Sandwich:
\small
\begin{equation*}
b_t =\exp\left(k\left(\sum_{j=1}^{d / 2} \cos \left(\frac{t}{r^{2 j / d}}\right)-\frac d 2 \right)\right), k > 0, r > 0.
\end{equation*}
 \normalsize

We first do the following transformations:
\small
\begin{equation}
\begin{aligned}
b_t &=\exp\left(k\left(\sum_{j=1}^{d / 2} \left(\cos \left(\frac{t}{r^{2 j / d}}\right)-1\right) \right)\right)\\
&=\prod_{j=1}^{d/2}\exp\left(k\left(\cos \left(\frac{t}{r^{2 j / d}}\right)-1\right)
\right),
\end{aligned}
\end{equation}
 \normalsize
then make a partition over $j$:
\small
\begin{equation}
\frac{t}{r^{2 j / d}} \ge \frac {\pi }2,
\end{equation}
 \normalsize
which is equivalent to:
\small
\begin{equation}
\begin{aligned}
2t &\ge \pi r^{2 j / d}, \frac{2t}{\pi}\ge r^{2j/d}, \\
\log_r\left(\frac{2t}{\pi} \right)&\ge 2j /d ,
j\le \frac{d \log_r\left(\frac{2t}{\pi} \right)}{2} \triangleq f(t).
\end{aligned}
\end{equation}
\normalsize
Therefore we have:
\small
\begin{equation}
\begin{aligned}
b_t
&=\prod_{1\le j\le f(t)}\exp\left(k\left(\cos \left(\frac{t}{r^{2 j / d}}\right)-1 \right)
\right)\\
&\times \prod_{f(t)< j \le d/2}\exp\left(k\left(\cos \left(\frac{t}{r^{2 j / d}}\right) -1\right)
\right).
\end{aligned}
\end{equation}
\normalsize
For the first part:
\small
\begin{equation}
\cos \left(\frac{t}{r^{2 j / d}}\right)-1< -1.
\end{equation}
\normalsize
For the second part:
\small
\begin{equation}
\cos \left(\frac{t}{r^{2 j / d}}\right)-1<0.
\end{equation}
\normalsize
Then:
\small
\begin{equation}
\begin{aligned}
\beta_t
&\le \prod_{1\le j\le f(t)} \exp(-k) \\
&= \exp(-k\lfloor f(t) \rfloor)\\
&\le \exp(k)\exp(-kf(t)) \\
&=\exp(k)\exp\left(- \frac{kd \log_r\left(\frac{2t}{\pi} \right)}{2} \right)
\triangleq g(t).
\end{aligned}
\end{equation}
\normalsize
According to Rabbe's test~\cite{knopp1956infinite}:
\small
\begin{equation}
\begin{aligned}
t\left(\frac{g(t)}{g({t+1})}-1\right)
&=t\left( \exp\left(\frac {kd} 2 \log_r \frac{2t+2}{2t} \right) -1\right)\\
&=t \left( \exp\left(\frac {kd} 2 \log_r \left(1+\frac 1 t \right) \right) -1\right)\\
&\sim t \left(\frac {kd} 2 \log_r \left(1+\frac 1 t \right) + O\left(\frac 1 {t^2} \right)\right)\\
&\sim t \left(\frac {kd} {2\ln r}\left(\frac 1 t -\frac 1 {2t^2}\right) +O\left(\frac 1 {t^2} \right)\right)\\
& \to \frac {kd}{2\ln r} ,
\end{aligned}
\end{equation}
\normalsize
if:
\small
\begin{equation}
\begin{aligned}
\frac{kd}{2\ln r} < 1, \quad
d < \frac{2\ln r}{k},
\end{aligned}
\end{equation}
\normalsize
then the series converges\footnote{Note that here we only show that the series corresponding to Sandwich is convergent under certain conditions. The upper bound here is relatively loose, and the conditions used in practice are broader.}.

\section{TRF example}
\label{trf-exmaple}
We use Alibi as an example to show the calculation of TRF. 
The $B_{ij}$ of Alibi can be written as:
\small
\begin{equation}
B_{ij}=\sum_{i=0}^{j - 1}\exp(-i)=\frac{1-\exp(-j)}{1-\exp(-1)},B =
\frac {1}{1-\exp(-1)}
\end{equation}
\normalsize
The TRF of Alibi can be calculated as:
\small
\begin{equation}
\small
\textstyle
\begin{aligned}
n_{\mathrm{the}}(\epsilon)
&=\inf_{j} \left (B_{ij}> B(1-\epsilon)  \right) \\
&=\inf_{j} \left (1-\exp(-j)> 1-\epsilon  \right) 
=\Theta(-\log\epsilon).
\end{aligned}
\end{equation}
 \normalsize
where $\Theta$ represents the upper and lower asymptotic bound.

\section{Configurations}
\begin{table*}[!ht]
\small
\center
\setlength{\tabcolsep}{0.6cm}
{

\begin{tabular}{l|l|l}
\hline\hline
Data   & WikiText-103   & Others       \\
\hline\hline
Decoder layers  & 6   & 12     \\
Hidden dimensions   & 512   & 768        \\
Number of heads     & 8    & 12       \\
FFN dimensions    & 2048    & 3072     \\
Tokenizer method & BPE    & BPE  \\
Src Vocab size & 50265  & 50265  \\
Sequence length   & 512  & 512\\
Total batch size & 128      & 128 \\
Number of updates/epochs    & 50k updates  & 50k updates \\
Warmup steps/epochs  & 4k steps   & 4k steps  \\
Peak learning rate    & 5e-4          & 5e-4    \\
Learning rate scheduler   & Inverse sqrt   & Inverse sqrt    \\
Optimizer  & Adam      & Adam      \\
Adam $\epsilon$    & 1e-8    & 1e-8    \\
Adam $(\beta_1,\beta_2)$     & (0.9, 0.98)    & (0.9, 0.98)  \\
Weight decay  & 0.01 & 0.01    \\
\hline\hline
\end{tabular}}
\caption{Detailed training configurations used in our experiments. ``Total batch size'' means $\mathrm{batch\_per\_gpu} \times \mathrm{update\_freq} \times \mathrm{num\_gpus}$.}
\label{configuration}
\end{table*}

\section{Pseudocode for TRF and ERF visualization}
\begin{lstlisting}
import torch

def draw(array, n=50):
    epsilon = torch.flip(torch.linspace(0, 1, n), dims=[0])
    index = torch.zeros(n)
    cusum = torch.sum(array)
    m = len(array)
    s = 0
    i = 0
    for j in range(m):
        eps = epsilon[i]
        while s >= cusum * (1 - eps) and i < n:
            index[i] = j
            if i < n - 1:
                i += 1
            else:
                break
            eps = epsilon[i]
        s += array[j]
    while i < n:
        index[i] = m
        i += 1
    
    return index / m, epsilon
\end{lstlisting}

\section{Detailed Experimental Results}
\begin{table}[!ht]
    \centering
    \begin{tabular}{llllllll}
    \hline \hline
    \multicolumn{8}{c}{\textbf{Wikitext-103}} \\ 
    \hline \hline
       Length & Sinusoidal & Alibi & Kerple-Log & Kerple-Power & Sandwich & Type1 & Type2 \\
        512 & 24.73 & 24.22 & 24.12 & 24.18 & 24.76 & 24.25 & 24.29 \\ 
        768 & 41.08 & 23.45 & 23.36 & 23.42 & 24.04 & 23.43 & 23.51 \\ 
        1024 & 62.71 & 23.06 & 22.93 & 22.98 & 23.63 & 23.03 & 23.09 \\ 
        1280 & 83.81 & 22.83 & 22.67 & 22.73 & 23.39 & 22.77 & 22.83 \\ 
        1536 & 102.28 & 22.66 & 22.45 & 22.54 & 23.21 & 22.55 & 22.63 \\
        1792 & 121.98 & 22.60 & 22.41 & 22.47 & 23.18 & 22.50 & 22.59 \\
        2048 & 138.17 & 22.52 & 22.28 & 22.37 & 23.08 & 22.38 & 22.48 \\
        3072 & 194.43 & 22.33 & 22.02 & 22.14 & 22.91 & 22.15 & 22.27 \\
        4096 & 259.55 & 22.26 & 21.97 & 22.08 & 22.96 & 22.09 & 22.21 \\
        5120 & 289.79 & 22.20 & 21.86 & 22.00 & 22.93 & 21.99 & 22.14 \\
        6144 & 337.46 & 22.17 & 21.87 & 21.96 & 23.06 & 21.97 & 22.11 \\
        7168 & 376.41 & 22.16 & 21.84 & 21.95 & 23.13 & 21.96 & 22.10 \\
        8192 & 406.95 & 22.14 & 21.82 & 21.94 & 23.20 & 21.95 & 22.08 \\
        9216 & 423.92 & 22.12 & 21.80 & 21.90 & 23.26 & 21.90 & 22.06 \\
   \hline \hline
    \multicolumn{8}{c}{\textbf{Books}} \\ 
    \hline \hline
        Length & Sinusoidal & Alibi & Kerple-Log & Kerple-Power & Sandwich & Type1 & Type2 \\
        512 & 7.49 & 7.28 & 7.34 & 7.31 & 7.64 & 7.30 & 7.35 \\ 
        768 & 10.43 & 7.15 & 7.21 & 7.18 & 7.55 & 7.18 & 7.22 \\ 
        1024 & 13.32 & 7.09 & 7.15 & 7.11 & 7.49 & 7.11 & 7.15 \\ 
        1280 & 15.53 & 7.06 & 7.11 & 7.08 & 7.47 & 7.08 & 7.12 \\
        1536 & 17.47 & 7.04 & 7.08 & 7.05 & 7.44 & 7.05 & 7.09 \\ 
        1792 & 19.02 & 7.03 & 7.06 & 7.03 & 7.42 & 7.03 & 7.06 \\ 
        2048 & 20.55 & 7.02 & 7.05 & 7.02 & 7.41 & 7.03 & 7.05 \\ 
        3072 & 24.70 & 7.00 & 7.02 & 7.00 & 7.38 & 7.00 & 7.03 \\
        4096 & 27.57 & 6.99 & 7.00 & 6.99 & 7.37 & 6.99 & 7.03 \\
        5120 & 29.54 & 6.99 & 7.00 & 6.99 & 7.36 & 6.99 & 7.03 \\
        6144 & 31.59 & 6.99 & 7.00 & 6.98 & 7.35 & 6.99 & 7.02 \\ 
        7168 & 32.41 & 6.98 & 7.00 & 6.98 & 7.35 & 6.99 & 7.02 \\ 
        8192 & 34.35 & 6.98 & 7.00 & 6.98 & 7.35 & 6.99 & 7.02 \\ 
        9216 & 34.70 & 6.98 & 7.01 & 6.98 & 7.35 & 6.99 & 7.02 \\ 
      \hline \hline
    \end{tabular}
\caption{\textbf{Sufficiency validation on Wikitext-103 and Books datasets.} To test the length extrapolation capability, we lengthen the inference sequence from 512 to 9216 and compute PPLs of our proposed Type 1 and Type 2 RPEs, as well as Alibi, Kerple, and Sandwich. All these methods are stable in PPL. For methods that cannot extrapolate, \eg~ Sinusoidal, its PPL grows rapidly.}
\label{table:result1}
\end{table}

\begin{table}[!ht]
    \centering
    \begin{tabular}{llllllll}
    \hline \hline
    \multicolumn{8}{c}{\textbf{Github}} \\ 
    \hline \hline
    Length & Sinusoidal & Alibi & Kerple-Log & Kerple-Power & Sandwich & Type1 & Type2 \\ 
        512 & 2.29 & 2.25 & 2.24 & 2.25 & 2.29 & 2.25 & 2.25 \\ 
        768 & 3.98 & 2.16 & 2.16 & 2.16 & 2.20 & 2.16 & 2.16 \\ 
        1024 & 7.91 & 2.12 & 2.12 & 2.12 & 2.16 & 2.12 & 2.12 \\ 
        1280 & 12.97 & 2.10 & 2.09 & 2.09 & 2.14 & 2.09 & 2.09 \\ 
        1536 & 18.66 & 2.09 & 2.07 & 2.08 & 2.12 & 2.08 & 2.08 \\
        1792 & 24.08 & 2.08 & 2.06 & 2.07 & 2.11 & 2.07 & 2.07 \\ 
        2048 & 30.02 & 2.07 & 2.05 & 2.07 & 2.10 & 2.06 & 2.06 \\ 
        3072 & 51.64 & 2.06 & 2.03 & 2.05 & 2.08 & 2.04 & 2.04 \\
        4096 & 70.62 & 2.05 & 2.02 & 2.04 & 2.07 & 2.03 & 2.03 \\ 
        5120 & 89.78 & 2.05 & 2.02 & 2.04 & 2.07 & 2.03 & 2.03 \\ 
        6144 & 101.28 & 2.05 & 2.01 & 2.04 & 2.06 & 2.03 & 2.03 \\ 
        7168 & 117.21 & 2.05 & 2.01 & 2.04 & 2.06 & 2.03 & 2.03 \\ 
        8192 & 130.15 & 2.05 & 2.01 & 2.03 & 2.06 & 2.02 & 2.03 \\ 
        9216 & 143.17 & 2.05 & 2.01 & 2.03 & 2.05 & 2.02 & 2.03 \\
      \hline \hline
    \multicolumn{8}{c}{\textbf{Wikibook}} \\ 
    \hline \hline
    Length & Sinusoidal & Alibi & Kerple-Log & Kerple-Power & Sandwich & Type1 & Type2 \\ 
        512 & 17.98 & 17.64 & 17.65 & 17.62 & 18.21 & 17.68 & 17.70 \\ 
        768 & 29.66 & 17.04 & 17.03 & 17.03 & 17.63 & 17.07 & 17.08 \\ 
        1024 & 47.31 & 16.76 & 16.73 & 16.73 & 17.35 & 16.76 & 16.80 \\ 
        1280 & 65.13 & 16.62 & 16.54 & 16.56 & 17.16 & 16.58 & 16.61 \\ 
        1536 & 83.16 & 16.51 & 16.41 & 16.44 & 17.04 & 16.44 & 16.49 \\ 
        1792 & 100.46 & 16.49 & 16.33 & 16.41 & 16.95 & 16.39 & 16.44 \\
        2048 & 116.94 & 16.43 & 16.26 & 16.36 & 16.89 & 16.31 & 16.39 \\
        3072 & 172.09 & 16.39 & 16.13 & 16.31 & 16.75 & 16.23 & 16.33 \\
        4096 & 231.86 & 16.39 & 16.14 & 16.29 & 16.73 & 16.24 & 16.36 \\
        5120 & 277.59 & 16.35 & 16.10 & 16.26 & 16.68 & 16.21 & 16.32 \\
        6144 & 312.17 & 16.35 & 16.12 & 16.25 & 16.66 & 16.22 & 16.32 \\
        7168 & 349.08 & 16.34 & 16.19 & 16.24 & 16.71 & 16.26 & 16.33 \\
        8192 & 390.81 & 16.35 & 16.22 & 16.25 & 16.71 & 16.30 & 16.36 \\
        9216 & 412.06 & 16.33 & 16.23 & 16.24 & 16.70 & 16.28 & 16.33 \\
         \hline \hline
    \end{tabular}
\caption{\textbf{Sufficiency validation on Github, WikiBook datasets.} To test the length extrapolation capability, we lengthen the inference sequence from 512 to 9216 and compute PPLs of our proposed Type 1 and Type 2 RPEs, as well as Alibi, Kerple, and Sandwich. All these methods are stable in PPL. For methods that cannot extrapolate, \eg~ Sinusoidal, its PPL grows rapidly.}
\label{table:result2}
\end{table}

\begin{table}[!ht]
    \centering
    \begin{tabular}{lllllllll}
    \hline \hline
  ~ & \multicolumn{2}{c}{\textbf{Wikitext-103}} 
    & \multicolumn{2}{c}{\textbf{Books}} 
    & \multicolumn{2}{c}{\textbf{Github}} 
    & \multicolumn{2}{c}{\textbf{Wikibook}}  \\
        \hline
        \hline
       Length & $\frac 1 n $ & $\frac 1 {n\ln n} $ & $\frac 1 n $ & $\frac 1 {n\ln n} $ & $\frac 1 n $ & $\frac 1 {n\ln n} $ & $\frac 1 n $ & $\frac 1 {n\ln n} $ \\ 
        512 & 24.67 & 24.64 & 7.40 & 7.35 & 2.28 & 2.27 & 17.91 & 17.90 \\ 
        768 & 23.87 & 23.81 & 7.28 & 7.24 & 2.19 & 2.18 & 17.28 & 17.28 \\ 
        1024 & 23.53 & 23.44 & 7.27 & 7.19 & 2.17 & 2.15 & 17.21 & 17.12 \\ 
        1280 & 23.50 & 23.25 & 7.41 & 7.20 & 2.19 & 2.15 & 17.70 & 17.28 \\ 
        1536 & 23.66 & 23.13 & 7.65 & 7.24 & 2.31 & 2.19 & 18.86 & 17.72 \\ 
        1792 & 24.20 & 23.22 & 7.97 & 7.30 & 2.52 & 2.28 & 20.74 & 18.65 \\ 
        2048 & 24.80 & 23.26 & 8.34 & 7.39 & 2.85 & 2.39 & 23.31 & 19.82 \\ 
        3072 & 28.31 & 23.65 & 9.97 & 7.82 & 5.05 & 3.15 & 38.46 & 27.94 \\ 
        4096 & 33.18 & 24.41 & 11.84 & 8.29 & 9.13 & 4.35 & 59.51 & 40.06 \\ 
        5120 & 37.65 & 25.01 & 14.15 & 8.80 & 15.79 & 5.93 & 84.60 & 55.12 \\ 
        6144 & 43.20 & 25.80 & 16.64 & 9.27 & 25.00 & 8.03 & 112.30 & 71.08 \\
        7168 & 48.07 & 26.39 & 18.99 & 9.73 & 36.01 & 10.26 & 140.34 & 89.79 \\
        8192 & 52.85 & 27.00 & 22.06 & 10.06 & 49.20 & 12.80 & 173.55 & 108.85 \\
        9216 & 58.46 & 27.63 & 26.34 & 10.65 & 76.78 & 16.35 & 204.10 & 133.36 \\ \hline  \hline
    \end{tabular}
\caption{\textbf{Necessity validation on Wikitext-103, Books, Github, WikiBook datasets.} We select two RPEs that do not satisfying Theorem~\ref{rpe-extrap-simp}, \eg~$b_n=\frac{1}{n}$ and $b_n=\frac{1}{n\ln n}$. We increase the inference sequence length from 512 to 9216 and compute the testing PPLs of them.
        Their PPLs increase rapidly as the inference sequence lengthens.
        }
        \label{table:result3}
\end{table}

\section{Heatmap}
\begin{figure*}[t]
 \tabcolsep=0.03cm
\begin{tabular}{ccc}
\includegraphics[width=0.33\linewidth]{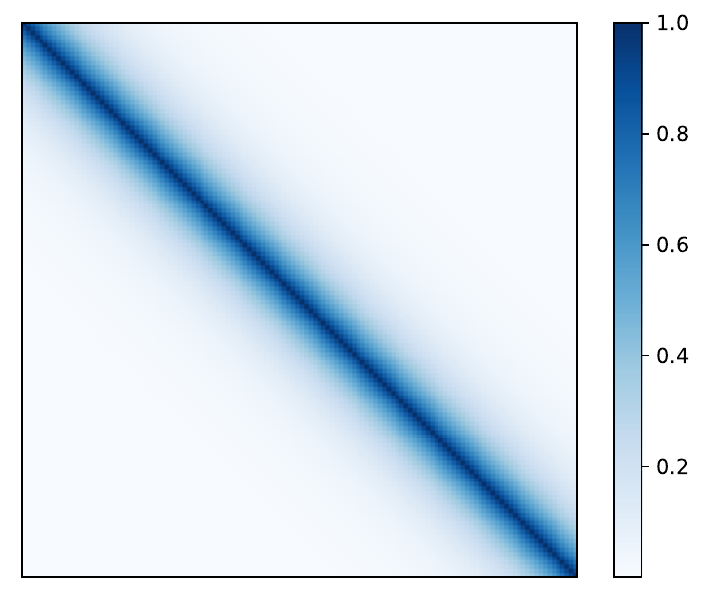} &
\includegraphics[width=0.33\linewidth]{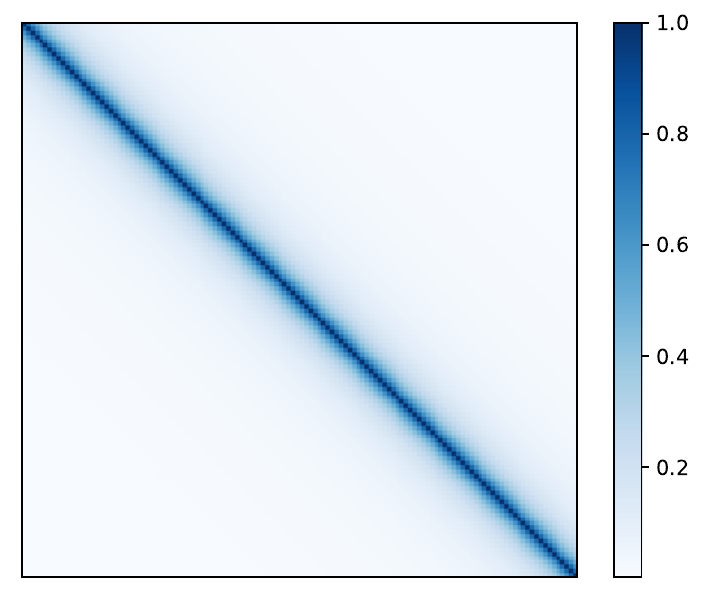} &
\includegraphics[width=0.33\linewidth]{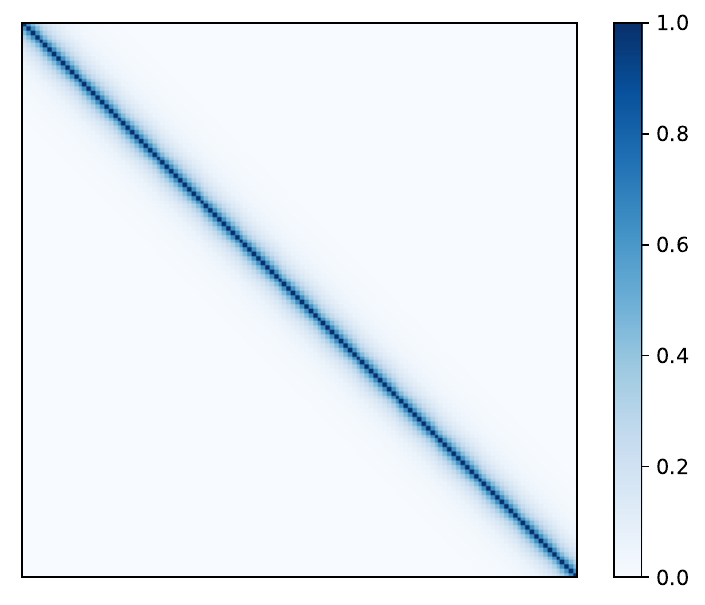} \\
(a) Alibi & (b) Kerple-Log & (c) Kerple-Power \\
\includegraphics[width=0.33\linewidth]{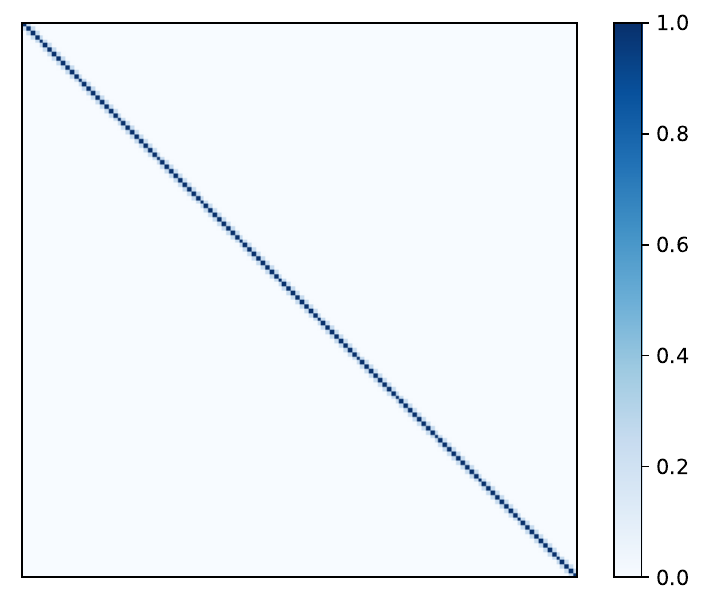}
 &
\includegraphics[width=0.33\linewidth]{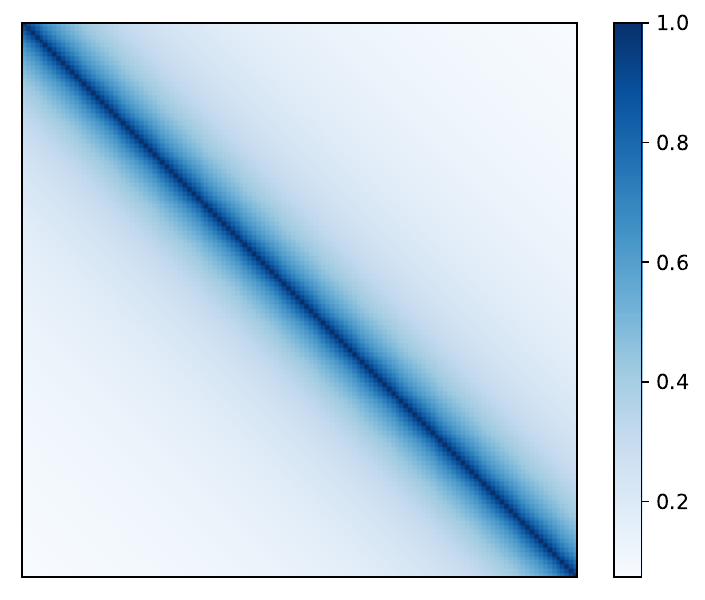}
 &
\includegraphics[width=0.33\linewidth]{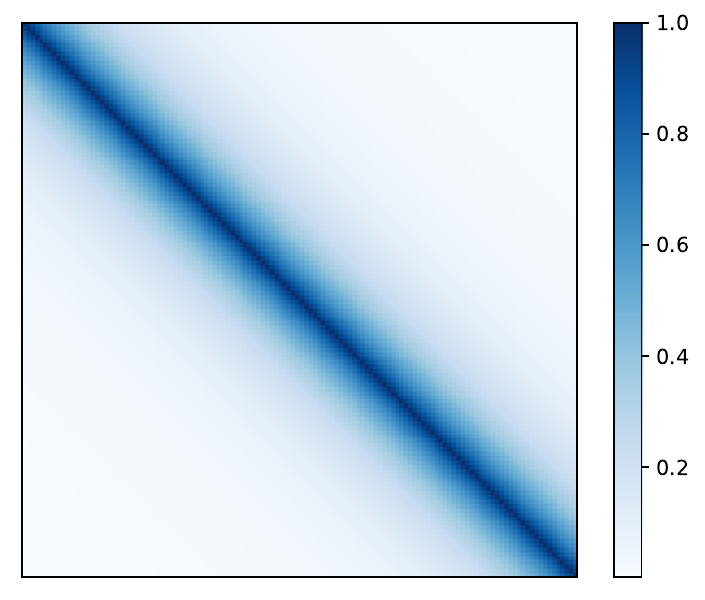}
\\
(d) Sandwich &  (e) $\frac 1 n $ & (f) $\frac 1 {n\ln n} $
\end{tabular}
\vspace{-3mm}
        \caption{\textbf{Visualization of RPE.} We additionally plot the heatmap of $\exp(\text{RPE})$ for all methods. Note that all methods are concentrated near the diagonal, including $\frac 1 n , \frac 1 {n\ln n}$, which shows that the concentration is not enough to guarantee extrapolation.}
        \label{fig:heatmapall}
        \vspace{-3mm}
\end{figure*}